\documentclass[11pt]{article}
 
\usepackage[top=1in,bottom=1in,left=1in,right=1in]{geometry}
\usepackage[numbers]{natbib}

\usepackage{amsmath,amssymb,amsfonts}
\usepackage{graphicx,color}
\usepackage{url}
\usepackage{booktabs}
\usepackage{mathtools}
\usepackage{microtype}
\usepackage{algorithm}
\usepackage{algorithmicx}
\usepackage[english]{babel}

\usepackage[toc,page,title,titletoc]{appendix}

\usepackage[dvipsnames]{xcolor}
\usepackage{tikz, pgfplots, pgfplotstable, subcaption}
\usepackage{hyperref}       
\hypersetup{
  colorlinks=true,
  linkcolor=black,
  urlcolor=black,
  citecolor=black
}
\usepackage[noend]{algpseudocode}

\usepackage{yhmath}

\usepackage{cleveref}
\crefname{section}{Section}{Sections}
\crefname{figure}{Figure}{Figures}
\crefname{mythm}{Theorem}{Theorems}
\crefname{mylem}{Lemma}{Lemmas}
\crefname{myrem}{Remark}{Remarks}
\crefname{myasp}{Assumption}{Assumptions}
\crefname{myrem}{Remark}{Remarks}
\crefname{appendix}{Appendix}{Appendicies}
\crefname{myprop}{Proposition}{Propositions}
\crefname{equation}{Equation}{Equations}
\crefname{table}{Table}{Tables}
\crefname{algorithm}{Algorithm}{Algorithm}

\usetikzlibrary{arrows.meta, calc, topaths, positioning, automata}
\pgfplotsset{compat=1.16}
\def\addlegendimage{\csname pgfplots@addlegendimage\endcsname}
\def\BibTeX{{\rm B\kern-.05em{\sc i\kern-.025em b}\kern-.08em
    T\kern-.1667em\lower.7ex\hbox{E}\kern-.125emX}}

\newcommand{\prob}[1]{\mathbb{P}( #1 )}
\newcommand{\Prob}[1]{\mathbb{P}\left( #1 \right)}
\newcommand{\mean}[1]{\mathbb{E} [ #1 ]}

\newcommand{\norm}[1]{ \|   #1  \| }
\newcommand{\ind}[1]{ \mathbf{I} \left\{#1 \right\} }
\newcommand{\argmax}[0]{ \mathop{\arg\max}  }
\newcommand{\argmin}[0]{ \mathop{\arg\min}  }

\newcommand{\meanX}[1]{\mathbb{E}_{X} [ #1 ]}
\newcommand{\meanYZ}[1]{\mathbb{E}_{YZ} [ #1 ]}
\newcommand{\meanXYZ}[1]{\mathbb{E}_{XYZ} [ #1 ]}
\newcommand{\normX}[1]{ \|   #1 \|_{L^2(\mathbb{P}_X)} }
\newcommand{\normYZ}[1]{ \|   #1 \|_{L^2(\mathbb{P}_{YZ})} }
\newcommand{\normH}[1]{ \|   #1 \|_{ \mathcal{H}  } }

\newcommand{\innerH}[2]{ \langle  #1, #2  \rangle_\mathcal{H} }
\newcommand{\innerU}[2]{ \langle  #1, #2  \rangle_\mathcal{U} }
\newcommand{\covYZ}[0]{ \mathfrak{C}_{YZ}  }
\newcommand{\covXYZ}[0]{ \mathfrak{C}_{XYZ}  }
\newcommand{\covYZX}[0]{ \mathfrak{C}_{YZX}  }

\newcommand{\KL}[0]{ \mathcal{K}\mathcal{L}  }

\newtheorem{mythm}{\bf{Theorem}}[section]
\newtheorem{mylem}{\bf{Lemma}}[section]
\newtheorem{myprop}{\bf{Proposition}}[section]
\newtheorem{mydef}{\bf{Definition}}[section]
\newtheorem{mycol}{\bf{Corollary}}[section]
\newtheorem{myrem}{\bf{Remark}}[section]
\newtheorem{myasp}{\bf{Assumption}}[section]
\newenvironment{myproof}{ {\noindent\it Proof.\ }}{\hfill $\square$\par}

\title{
  Dual Instrumental Method for Confounded Kernelized Bandits
}

\usepackage{authblk}
\author[*]{Xueping Gong}
\author[$\dagger$]{Jiheng Zhang}

\affil[*$\dagger$]{Department of Industrial Engineering and Decision Analytics}
\affil[*$\dagger$]{The Hong Kong University of Science and Technology}

\date{}

\begin{document}
\maketitle

\begin{abstract}
  The contextual bandit problem is a theoretically justified framework with wide applications in various fields.
  While the previous study on this problem usually requires independence between noise and contexts,
  our work considers a more sensible setting where the noise becomes a latent confounder that affects both contexts and rewards.
  Such a confounded setting is more realistic and could expand to a broader range of applications.
  However, the unresolved confounder will cause a bias in reward function estimation and thus lead to a large regret.
  To deal with the challenges brought by the confounder, 
  we apply the dual instrumental variable regression, which can correctly identify the true reward function. 
  We prove the convergence rate of this method is near-optimal in two types of widely used reproducing kernel Hilbert spaces. 
  Therefore, we can design computationally efficient and regret-optimal algorithms based on the theoretical guarantees for confounded bandit problems.
  The numerical results illustrate the efficacy of our proposed algorithms in the confounded bandit setting.
\end{abstract}

\section{Introduction}

Contextual bandit problems have been studied to capture the trade-off between \emph{exploration} and \emph{exploitation} in online decision-making.
Various formulations of the problems show wide applications ranging from scheduling, dynamic pricing, packet routing, online auctions, e-commerce, and matching markets \citep{MLsurvey}.
At each round, a learner chooses an \emph{action} for an observed \emph{context} to generate a \emph{reward}, which depends on the action and context. 
The goal is to maximize the cumulative expected rewards, or equivalently, to minimize the cumulative expected \emph{regret}. 
Many algorithms have been designed in literature (see a book of bandit algorithm summary \citep{banditBook}) to achieve near-optimal regret rates. 

Many existing studies on contextual bandits, including linear bandits \citep{improvedUCB},
generalized linear bandits \citep{GLbandit} and kernelized bandits \citep{kernelisedContextualBandits},
rely on one essential assumption: \emph{the independence between noise and contexts}.
In this paper, we relax this assumption by modeling the correlation using causal graphs where the noise becomes a latent confounder. 
Such a causal relationship is arguably sensible for practical applications in the real world. 
Many practical problems can be modeled using this framework \citep{CounterfactualPrediction,DeepProxy,dualIV}.
Under such a framework, we need to estimate the unknown function from noisy and possibly high-dimensional samples affected by the unobserved confounders while striking a balance between exploration and exploitation to achieve optimal regret.
We apply causal tools, e.g., the instrumental variable (IV) regression, to tackle the challenge brought by latent confounders. 
Combined with the kernel trick and dual formulation, the instrumental variable method elegantly does regressions in reproducing kernel Hilbert spaces (RKHS) and accurately identify the causal effect. 
To deal with the non-i.i.d.\ issue of bandit data, we divide the time horizon into epochs, in each of which, our proposed action sampling policy can effectively balance the exploration and exploitation and efficiently reduce the computational burden.

\paragraph{Contributions.}
First, our work generalizes the kernelized contextual bandit in a causal setting with latent confounders.
In this way, we allow the noise to become confounders that affect both contexts and rewards compared with previous works \citep{improvedUCB,MaternKerbelBandit,kernelisedContextualBandits,EfficientKernelUCB}.
We show that in such confounded settings, the learner can still achieve a near-optimal regret (up to $\mathcal{O}(\log\log T)$ terms) by our \cref{alg: DIV-BLS}, 
which is comparable performance with existing bandit algorithms in unconfounded settings. 
Our algorithm is computationally efficient because we reduce the number of solving optimization problems from $\mathcal{O}(T)$ to $\mathcal{O}(\log T)$ (or $\mathcal{O}(\log\log T)$ if $T$ is known) by an epoch-based learning strategy.
Second, we analyze the convergence rate of the dual IV method and give a guideline on choosing the regularization parameters, which may be of independent interest. 
In \citep{dualIV}, only the consistency of the dual IV estimator is proved under realizability, invertibility, and continuity assumptions.
We consider the cases of both finite-dimensional (\cref{thm: oracle inequality for finite rank}) and infinite-dimensional (\cref{thm: oracle inequality}) spaces, 
and prove that this method achieves optimal convergence rates with large probability under the same conditions (\cref{thm: lower bound of finite dimensional spaces},\cref{thm: lower bound}).

\subsection{Related Works}

\emph{Unobserved Confounders.}
The study of unobserved confounders is one of the central themes in the modern literature of causal inference \citep{causaloverview,causalsurvey}. 
With the presence of unobserved confounders, many novel methods are proposed; see \citep{causaloverview,causalsurvey} for an overview.
These methods are also widely studied in bandit settings. 
\citet{bandits_causal_approaches} point out the possibility and necessity of causal approaches in bandits when faced with confounders. 
A novel causal bandit model is proposed in \cite{causalbandit} to illustrate the causal relationships of actions by causal graphs.
Further, \citet{CausalBanditsLearning} combine the causal methods with traditional bandit algorithms to demonstrate that causal approaches can significantly improve the regret bounds. 

\emph{Instrumental Variable Regression.}
IV regression is another method for learning causal relationships with observational data. 
When measurements of input and output are confounded, the causal relationship, also called the structural relationship, 
can be identified if an instrumental variable is available (see \cref{fig: causal model with instrumental variables}). 
Instrumental variable regression involves two-stage regression (2SLS), and \citet{kernelIV,proximalkernel} generalize this method to nonlinear settings. 
They also provide consistency guarantees for their kernel instrumental variable algorithm. 
The idea of dual formulation simplifies traditional two-stage methods. 
Several works \cite{dualIV,BayesianKernelMeanEmbeddings} propose such ideas with similar mathematical structure.

\emph{Instrumental Variable in Bandits.}
A few papers apply IV regression in machine learning. 
\citet{IAB} formalize an instrumented-armed bandit (IAB) framework in a multi-armed bandit setting.
The arm pull is the choice of the instrumental variable which influences rather than guarantees the application of the treatment. 
As an application of IAB, \citet{RecommendationInstrument} develop a novel recommendation mechanism that views the recommendation as a form of instrumental variables. 
Such mechanisms strategically select instruments to incentive compliance over time to achieve optimal regrets up to logarithmic terms. 
However, both works assume the linearity of structural equations.
When structural functions are nonlinear, \citet{DeepProxy} show that IV implementation can be broken into two supervised stages, which can be targeted with deep networks.
\citet{DIV_NN} take an important step in this direction and provide convergence analysis for neural networks. 

\emph{Kernelized Bandit.}
The kernelized bandit was originally formulated by \citet{GPbandit}. 
This work generalizes stochastic linear optimization in a bandit setting, 
where the unknown reward function comes from a finite-dimensional reproducing kernel Hilbert space (RKHS). 
The smoothness assumptions about functions are encoded through the choice of kernels in a flexible nonparametric fashion.
\citet{GPbandit} resolve the problem of deriving regret bounds via GP optimization in RKHS. 
\citet{kernelisedContextualBandits} propose Kernel-UCB algorithm and obtain $\tilde{\mathcal{O}} ( \sqrt{ \tilde{d} T} ) $ regret, where $\tilde{d}$ is the effective dimension of data. 
\citet{MaternKerbelBandit} consider the cases when the kernel is of infinite rank, e.g., Mat{\'e}rn kernel. 
They use the improved GP-UCB algorithm to achieve a suboptimal regret upper bound $\tilde{\mathcal{O}} (T^{\frac{d(d+1)}{d(d+1) +2\nu }})$, where $\nu$ captures the smoothness of Mat{\'e}rn kernels and $d$ is the dimension of contexts.
Later, \citet{EfficientKernelUCB} improve computational efficiency of kernelized UCB algorithms. 
They also show that the concepts of information gain in \cite{GPbandit} and effective dimension in \cite{kernelisedContextualBandits,EfficientKernelUCB} are equivalent up to logarithmic factors.

\paragraph{Overview.}
In \cref{sec: problem formulation}, we propose a contextual bandit with a nonlinear reward function and latent confounders. 
We also introduce the instrumental variable regression in this section, 
in order to tackle the challenge brought by the unobserved confounders. 
In \cref{sec: methods}, we propose the dual method to perform IV regression efficiently in the bandit setting. 
Then we analyze the dual method in reproducing kernel Hilbert spaces 
and obtain the concentration inequality. 
Next, we design a bandit algorithm that combines the idea of dual IV regression and epoch learning strategy. 
We show the regret upper and lower bounds in \cref{sec: regret analysis}. 
Moreover, we illustrate the numerical results in \cref{sec: numerical experiments}.
In appendices, we provide all the proofs,  
and discuss the concentration inequalities and the regret of the new bandit algorithm for infinite-dimensional RKHSs.

\paragraph{Notations.} We define the following notations which will be used throughout the paper. 
We denote $\innerH{\cdot}{\cdot}$ and $\normH{\cdot}$ as the inner product of the Hilbert space $\mathcal{H}$ and its induced norm, respectively. 
The $L^2$-norm of function $f$ associated with random variable $X$ is defined as $\normX{f}: = \sqrt{ \meanX{f^2(X)}}$.
For function $f,g$, we write $f(x) = \mathcal{O} (g(x))$, 
if there exists a constant $C>0$ such that $f(x)\leq Cg(x)$, 
and write $f(x) = \Omega (g(x))$ if $g(x)=\mathcal{O} (f(x))$.
The notation $\simeq$ means two quantities are of the same order up to a constant. 
We denote $\otimes$ as the tensor product.

\section{Problem Formulation}
\label{sec: problem formulation}

\subsection{Contextual Bandit With A Latent Confounder}
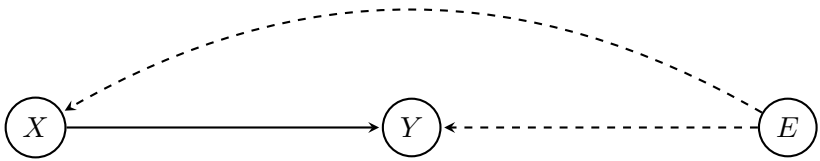
\begin{figure}[h]
  \center
    \begin{tikzpicture}[->,shorten >=1pt,auto,node distance=5cm,-stealth,
        thick,base node/.style={circle,draw,minimum size=48pt}, real node/.style={double,circle,draw,minimum size=50pt},scale = 1]
        
        \node[shape=circle,draw=black](1){$X$};
        \node[shape=circle,draw=black](2)[right of = 1 ]{$Y$};
        \node[shape=circle,draw=black](0)[right of = 2]{$E$};
  
        \path[]
        (1) edge node { } (2)
        (0) edge [dashed, bend right] node  {  } (1)
        (0) edge [dashed, right] node { } (2);
              
    \end{tikzpicture}
    \caption{A causal model of $X$ and $Y$ with unobserved confounder $E$. }
    \label{fig: causal model of correlated bandit}
\end{figure}

We consider a contextual bandit problem with $K$ arms. 
A learner interacts with the environment in several rounds $t=1,2,\cdots,T$, 
where $T$ is the time horizon. 
At each round $t$, the environment generates a context $c_t$ in a compact set $\mathcal{C}$. 
The learner is given an action set $\mathcal{A}$ and chooses an action $a_t\in\mathcal{A}$.
Note that $\mathcal{A}$ may change over time, but the size of $\mathcal{A}$ keeps fixed, i.e., $|\mathcal{A}|=K$.
The context and action spaces can be discrete or included in $ \mathbb{R}^d$. 
The learner will obtain a reward $y_t$ after it chooses the action $a_t$.
We model the reward function for the context-action pair as 
\begin{equation}
  \label{eq: reward function}
y_{t}=f(c_t,a_t)+e_t,
\end{equation}
where $e_t$ is a bounded $\sigma^2$-subgaussian noise term. 
Furthermore, we consider a correlated bandit problem
and allow for noise $e_t$ that is potentially correlated with the context $c_t$, namely, 
$$
\mathbb{E}[e_t|c_t]\neq 0.
$$
For notation brevity, we let $x_t := (c_t,a_t)$ and $\mathcal{X} := \mathcal{C}\times \mathcal{A}$.
We use a causal model in \cref{fig: causal model of correlated bandit} to describe the correlation.
As the structural causal graph shows, 
the noise term $E$ serves as an unobserved confounder, making the causal identification between the context-action pair $X$ and the reward $Y$ challenging.

The natural filtration $\mathcal{F}_t$ is defined w.r.t.\ the sequence of contexts and the collected rewards up to $t$.
A policy $\pi$ of the learner is a non-anticipatory decision sequence of actions in $\mathcal{A}$,
i.e., $\pi_t : \mathcal{F}_{t-1} \to \mathcal{A}$.
The expected regret of an algorithm is defined to be 
$$
Reg(T) = \sum_{t=1}^T \mathbb{E}[ f(c_t,a_t^*) - f(c_t,a_t) ],
$$
where the optimal action is
$$
a_t^*:=\argmax_{ a \in \mathcal{A}} f(c_t,a), 
$$
and $a_t =\pi(\mathcal{F}_{t-1})$ is the arm pulled according to the policy $\pi$ at step $t$.
Our goal is to find algorithms to minimize the above regret. 

Our proposed contextual bandit framework is a generalization of kernelized contextual bandit problems \cite{improvedUCB, kernelisedContextualBandits,EfficientKernelUCB} in a causal setting. 
The following are special cases of our model: 
\begin{itemize}
\item When $f$ is independent of the context, i.e., $f(c,a) = g(a)$ for some function $g$, our framework can be reduced to multi-armed bandit problems. 
\item When $f$ is linear in contexts and actions, i.e., $f(c,a) = a^\top c$, our proposed bandit problems are linear contextual bandits with latent confounders. 
\item When the hidden confounder is independent of context, these bandit problems are kernelized contextual bandits.
\end{itemize}
Our proposed framework involves three main challenges: 
\begin{enumerate}
  \item deal with the effect of latent confounders and obtain an unbiased and accurate estimator of the reward function;
  \item perform efficient regressions on the reward function with controlled non-linearity; 
  \item balance the trade-off between exploration and exploitation to achieve optimal regret.
\end{enumerate}
We address the first challenge by dual IV regressions, the second by a kernel trick, 
and the third by an epoch learning strategy and an action sampling policy.

\subsection{Instrumental Variable Regression}

In this section, we introduce instrumental variable (IV) regression. 
Direct regression of $Y$ on $X$ will lead to estimation bias, 
as the context components of $X$ are confounded by the noise variable $E$. 
The causality between the confounder variable $E$ and the context $X$ makes the conditional expectation $\mean{e_t|x_t}$ nonzero, 
thus
$$
\mean{Y|X} = f(X) + \mean{E|X} 
$$
is not an unbiased estimator for $f$. 
If samples of the confounder are available, the causal effect can be identified by backdoor adjustment \cite{causaloverview}, 
but this technique will fail when the confounder is hidden.
Fortunately, the presence of \emph{instrumental variables} allows us to estimate an unbiased $\hat{f}(x)$ that captures the structural relationship between $X$ and $Y$. 
These are sets of variables $Z$ that satisfy the following three conditions \cite{CounterfactualPrediction}.

\begin{enumerate}
  \item[] \emph{Relevance}: the conditional distribution $\prob{x|z}$ of $X$ given $Z=z$ is not constant in $z$.

  \item[] \emph{Exclusion}: $Z$ does not play a role in the expectation of \eqref{eq: reward function}, i.e., $Z\bot Y| (X, E)$.
  
  \item[] \emph{Unconfoundedness}: $Z$ is independent of the confounder $E$, i.e., $Z\bot E$.
\end{enumerate}

\begin{figure}[h]
  \center
    \begin{tikzpicture}[->,shorten >=1pt,auto,node distance=3cm,-stealth,
        thick,base node/.style={circle,draw,minimum size=48pt}, real node/.style={double,circle,draw,minimum size=50pt},scale = 0.8]
        
        \node[shape=circle,draw=black](3){$Z$};
        \node[shape=circle,draw=black](1)[right of = 3]{$X$};
        \node[shape=circle,draw=black](2)[right of = 1 ]{$Y$};
        \node[shape=circle,draw=black](0)[right of = 2]{$E$};

        \path[]
        (1) edge node { } (2)
        (0) edge [dashed,bend right]node { } (1)
        (3) edge node { } (1)
        (0) edge [dashed,right] node { } (2);
              
    \end{tikzpicture}
    \caption{A causal model with unobserved confounders and instrumental variables. }
    \label{fig: causal model with instrumental variables}
\end{figure}
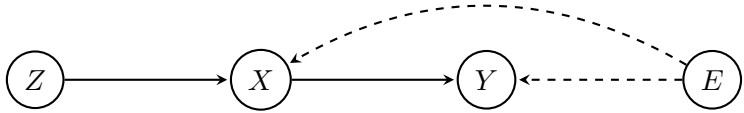

\begin{myrem}
  The causal graph (\cref{fig: causal model with instrumental variables}) with instrumental variables illustrates these conditions clearly. 
  The relevance condition means $Z$ should have a direct causal effect on $X$. 
  The instrumental variables $Z$, also called proxy variables, should reflect some information of $X$. 
  The exclusion condition requires that given $X$ and $E$, $Y$ and $Z$ are independent. 
  This condition indicates that $Z$ should not have a direct causal effect on $Y$; otherwise, $Z$ would bring a new confounded factor into $X$ and $Y$. 
  The unconfoundedness condition demonstrates that $Z$ is not affected by unobserved confounder $E$, 
  so we can utilize this property to elegantly eliminate the influences of $E$ on estimation bias.
  Under the additive error assumption made in \eqref{eq: reward function}, 
  unconfoundedness of the instrumental variables is unnecessary: we could replace this assumption with the weaker mean independence assumption $\mean{E|Z} = 0 $ without changing anything. 
\end{myrem}

As \cref{fig: causal model with instrumental variables} depicts, 
a variation in $X$ comes from both $E$ and $Z$. 
Intuitively speaking, the external source of variation from $Z$ can help improve an estimation by removing the effect of $E$ on $X$.
The followings are two examples concerning instrumental variables in literature.
(1) In an air-travel demand example in \cite{CounterfactualPrediction,DeepProxy}, 
$X,Y$ are the price and the sales, respectively. 
There is a big ``conference'' $E$, unobserved to the decision-maker, which drives demand and price. 
The instrumental variable $Z$ is the cost of fuel which influences sales only by price. 
(2) In a demand design example in \cite{dualIV,DeepProxy},
$X,Y$ are the price and the sales, respectively. 
The price and the sales are confounded by the customer sentiment $E$. 
The customer sentiment is observable to some degree but very hard to quantify.
The instrumental variable $Z$ is a supply cost shifter.

After introducing the instrumental variables, we can correctly identify the true structural function.
Taking expectation on both sides of \eqref{eq: reward function} conditioning on $Z$ yields the following equality:
\begin{equation}
  \label{eq: functional equation for f}
\begin{array}{rcll}
\mean{Y|Z} &=& \mean{f(X)|Z } + \mean{E |Z }    \\
&=& \mean{f(X)|Z }     \\
&=& \int_{\mathcal{X}}  f(x) d \prob{x|Z}. 
\end{array} 
\end{equation}
Under the relevance assumption, we can approximate this equation to obtain an unbiased estimator $\hat{f}$ for $f$, 
because $X,Y,Z$ are all observable. 
In the bandit setting, we assume that the learner can observe an extra instrumental variable $z_t$ at each round,
and that the tuple $(c_t,z_t)$ is i.i.d.\ generated according to \cref{fig: causal model with instrumental variables}.
At round $t$, the learner chooses an arm $a_t\in\mathcal{A}$ based on historical information $\mathcal{F}_{t-1}$, 
and observe a reward $y_t$.
We extend the natural filtration to $\mathcal{F}_t =\sigma( \{a_i,c_i,y_i,z_i\}_{i=1}^t  )$
by adding the information generated by the instrumental variable $Z$.

The consistent estimation of the structural functions involves two stages of nonparametric regressions. 
Many papers \cite{DeepProxy,CounterfactualPrediction,kernelIV} propose elegant and novel methods to perform nonparametric regression targeted at instrumental regressions.
However, they make strong smoothness assumptions on the structural functions \cite{CEbyIV:linear_f,RecommendationInstrument} and conditional probability \cite{IV_conditional_expectation},
or require sufficiently large dataset as their algorithms suffer from the curse of dimensionality \cite{DeepProxy,kernelIV}.

\section{Methods}
\label{sec: methods}

The availability of instrumental variable $Z$ makes the causal effect identifiable in our kernelized contextual bandit setting. 
IV regression is a useful tool and involves estimating conditional distribution $\prob{X|Z}$ or the conditional mean embedding $\mathbb{E}_{X|Z}[\cdot]$. 
However, besides the intractability of estimating such conditional quantities, 
the requirement of an exploration-exploitation trade-off makes the whole problem more challenging.
Since the action $a_t$ consists of the input parameter $X$ in bandit problems, 
estimating $\prob{X|Z}$ imposes a statistical challenge.
Moreover, the first-stage estimation creates a finite-sample bias in the second-stage estimation.
A sample-inefficient algorithm may induce suboptimal regret
because the regret is highly dependent on the convergence rates of reward functions.
However, these direct methods for general conditional probabilities are usually suboptimal,
especially when $X$ and $Z$ are high-dimensional. 
Furthermore, only one sample can be observed at each round in bandit problems.
To tackle the challenges brought by confounders in bandit settings, 
we need to find a suitable set of conditional probabilities to make the whole problem tractable and avoid directly estimating these quantities.
Then we design algorithms for confounded contextual bandit problems
based on the idea of dual instrumental variable regression and an epoch learning strategy to achieve optimal regrets.

\subsection{Dual Formulation}

In this subsection, we first provide a dual formulation of IV regressions. 
To avoid ambiguity, we denote $f^*$ as the truth and $f$ as a general functional variable.
Since samples are stochastic and noisy, we reformulate equation \eqref{eq: functional equation for f} as the following minimization problem
\begin{equation}
   \label{eq: minimization problem}
\min_{f\in\mathcal{H}} R(f) := \meanYZ{\ell(Y,\mathbb{E}_{X|Z}[f(X)]  ) },
\end{equation}
where $\ell(y,y')=\frac{1}{2}(y-y')^2$ is the quadratic loss. 
The true structural function $f^*$ can be identified by the optimum of the above minimization problem if $f^*$ is in the function space $\mathcal{H}$.
The conditional expectation operator $\mathbb{E}_{X|Z}[\cdot]$ is difficult to approximate by samples, 
because of the limited sample size and the possibly high dimensions of $X$ and $Z$.
Since \eqref{eq: minimization problem} is a convex problem with respect to $f$, 
we can solve its dual problem to obtain a solution. 
The loss $\ell_f=\frac{1}{2}(y-f(x))^2$ is convex and continuous, and its convex conjugate is $\ell^*_u = uy +\frac{1}{2}u^2$.
Hence, problem \eqref{eq: minimization problem} is equivalent to the following maximization problem
\begin{equation}
   R(f) = \max_{u\in\mathcal{U}} \Psi(f,u),
\end{equation}
where
$$
\begin{array}{rcl}
\Psi(f,u)   &=& \meanYZ{  \mathbb{E}_{X|Z}[f(X)-Y]u(Y,Z)  -\frac{1}{2}u(Y,Z)^2  } \\
     &=& \meanXYZ{  f(X)u(Y,Z) -  Yu(Y,Z) -\frac{1}{2}u(Y,Z)^2} \\
     &=& \meanXYZ{  (f(X)-Y)u(Y,Z)}  -\frac{1}{2}  \meanYZ{u(Y,Z)^2}.
\end{array}
$$
The first equality is due to the interchangeability of expectation and maximization \cite{dualIV}. 
The second equality is a result of total expectation property.
The function spaces $\mathcal{H}$ and $\mathcal{U}$ will be specified later.
It follows from \cref{prop: risk}, the optimal solution $u^*(y,z)$ to the above problem takes the form $u^*(y,z)=\mathbb{E}_{X|z}[f(X)]-y$.
This formulation is inspired by the mathematical resemblance of non-linear IV to two-stage problems in previous works \citep{stochasticprogramming,dualEmbedding,dualIV,BayesianKernelMeanEmbeddings}.

The dual instrumental variable regression has various advantages. 
Compared to previous work that involves density estimation and vector-valued regression in the first-stage regression, 
estimating real-value functions is arguably easier. 
Moreover, functions $u$ and $f$ can be estimated in the same round in the bandit setting, 
so the whole sample set can be fully utilized. 
Furthermore, we show that the convergence rate of the dual method is optimal up to a probability term in \cref{thm: oracle inequality for finite rank}.
Based on it, we can adaptively explore and exploit actions to achieve optimal regret.

\paragraph{RKHS.} We now discuss the selection of the function spaces. 
We choose $\mathcal{H}$ and $\mathcal{U}$ to be reproducing kernel Hilbert spaces associated with positive definite and continuous kernels $k:\mathcal{X}\times \mathcal{X} \to \mathbb{R}$ and $l:(\mathcal{Y}\times \mathcal{Z}) \times (\mathcal{Y}\times \mathcal{Z}) \to \mathbb{R}$, respectively.
Let $\phi(x):= k(x,\cdot)$ and $\varphi(y,z):= l((y,z),\cdot)$ be the canonical feature maps of $\mathcal{H}$ and $\mathcal{U}$, respectively.
Due to the properties of RKHS, $\Psi(f,u)$ can be rewritten as 
$$
\Psi(f,u) = \innerU{ \covYZX f -r}{u}  - \frac{1}{2} \innerU{u}{\covYZ u},
$$
where
$$
r = \mathbb{E}_{YZ}[Y\varphi(Y,Z)],
\covYZ = \mathbb{E}_{YZ}[ \varphi(Y,Z) \otimes \varphi(Y,Z)  ],
\covYZX = \mathbb{E}_{XYZ}[ \varphi(Y,Z) \otimes \phi(X)  ].
$$
For a deep insight into the covariance operator $\covYZ $ and the cross-covariance operator $\covYZX$ (its adjoint is denoted as $\covXYZ$), 
we refer the readers to \cref{subsec: RKHS}.

Since $\Psi(f,u)$ is quadratic in $u$, the maximizer $u^*$ satisfies 
\begin{equation}
   \label{eq: relationship between f and u}
\covYZ u^* =\covYZX f - r.
\end{equation}
Substituting $u^*$ back yields a closed-form solution:
\begin{equation}
   \label{eq: closed-form solution to f}
f^* = ( \covXYZ \covYZ^{-1} \covYZX )^{-1} \covXYZ \covYZ^{-1}r
\end{equation}
if the referred linear operators are invertible. 
This expression makes it easy to analyze in theory and estimate in practice. 
In empirical versions, corresponding operators might not be invertible, 
so a regularizer $\lambda =(\lambda_1,\lambda_2)$ can be added to the solution. 
The modified object will be 
$$
{\Psi}_\lambda(f,u) = \mathbb{E}_{XYZ}[(f(X)-Y)u(Y,Z)]+ \frac{\lambda_2}{2} \norm{f}_\mathcal{H}^2    -\frac{1}{2} \mathbb{E}_{YZ}[u(Y,Z)^2]-\frac{\lambda_1}{2} \norm{u}_\mathcal{U}^2 , \\
$$
and the empirical version is 
$$
\hat{\Psi}_\lambda(f,u) = \frac{1}{n} \sum_{i=1}^n (f(x_i)-y_i) u(y_i,z_i) + \frac{\lambda_2}{2} \norm{f}_\mathcal{H}^2    -\frac{1}{2n} \sum_{i=1}^n u(y_i,z_i)^2 - \frac{\lambda_1}{2} \norm{u}_\mathcal{U}^2    .
$$
The solution obtained from the following empirical minimization-maximization problem 
$$
\min_{f\in\mathcal{H}}\max_{u\in\mathcal{U}} \hat{\Psi}_\lambda(f,u)
$$
is denoted by dualIV($k,l,\lambda_1,\lambda_2$). 
This regression can be viewed as a generalized least squares regression in RKHSs, 
and a convenient formulation exists in numerical experiments (see \cref{alg: dualIV} in \cref{sec: dual method}).

Before moving on to the design of algorithms in kernelized contextual bandit settings, 
we summarize the assumptions for function spaces and kernels. 
\begin{myasp} [Realizability] 
   \label{asp: realizability}
The RKHSs $\mathcal{H}$ and $\mathcal{U}$ are correctly specified for the reward function and its dual function, 
i.e., $f^*\in \mathcal{H}$ and $u^* \in \mathcal{U}$.
\end{myasp}

The realizability assumption is standard in the literature \citep{kernelIV,dualIV,IV_conditional_expectation}.
This assumption indicates that the function spaces are ``large'' enough to contain the functions of our interest.
The realizability for $u^*$ also implicitly makes assumptions on the conditional distribution $\prob{X|Z}$.
Some works on bandit problems \cite{improvedUCB,kernelisedContextualBandits} make this assumption by restricting the rewards in a known and bounded interval.

\begin{myasp} [Invertibility] 
   \label{asp: invertibility}
   The covariance operators $\covYZ, \covXYZ \covYZ^{-1} \covYZX$ referred in \eqref{eq: closed-form solution to f} are invertible.
\end{myasp}

The invertibility assumption is sensible and applicable. 
Such (uncentered) covariance operators capture the covariance of two elements in RKHSs \cite{RKHStheory}. 
For example, these covariance operators are equivalent to covariance matrices in some sense when RKHSs are finite-dimensional. 
This assumption essentially requires that covariance matrices are positive definite, 
which is satisfied in many cases.

\begin{myasp} [Continuity]
   \label{asp: continuity}
   The referred kernels of $\mathcal{H}$ and $\mathcal{U}$ are continuous on compact sets.
\end{myasp}

The continuity of kernels indicates their boundedness.
A wide range of kernel families satisfies the continuity assumption, 
including linear kernels, polynomial kernels, RBF kernels, and Sobolev-type kernels.
Some works such as \cite{optRatesRLSR,RKHSminimax,svm} impose the continuity or directly suppose the boundedness of kernels as a weaker assumption.

\paragraph{Concentration Inequalities.}
An important step for bandit algorithm design is to establish a concentration inequality. 
The convergence rate of dual methods will provide a useful guide to the algorithm design.
However, the above assumptions are not enough to establish a concentration inequality 
due to the varieties of kernels and spaces. 
We classify the spaces by their dimensions because the Hilbert spaces of the same finite dimensions are isomorphic. 
We prove the following theorem for $\tilde{d}$-dimensional spaces.  
Mathematically, the kernels of such spaces can be expanded in terms of $\tilde{d}$ basis functions. 
Function classes of this type include linear functions and polynomial functions.
Generally, any function space with finite VC-dimension satisfies this condition \cite{RKHSminimax}. 

\begin{mythm}
   \label{thm: oracle inequality for finite rank}
   Let the $\tilde{d}$-dimensional RKHSs $\mathcal{H}$ and $\mathcal{U}$ associated with kernels $k$ and $l$ satisfying \cref{asp: realizability,asp: invertibility,asp: continuity}.
   Consider a dataset $(X_i,Y_i,Z_i)_{i=1}^n$ i.i.d.\ sampled according to \cref{fig: causal model with instrumental variables}, 
   and $\hat{f}_n$ is obtained from dualIV with regularization parameters $\lambda_i \simeq  \sqrt{ \frac{  \tilde{d} \tau   }{n} }  $ for $i=1,2$.
   Then, there exists a constant $M$ 
   which depends on the true structural function $f^*$ and spaces $\mathcal{H},\mathcal{U}$,
   such that for all $\tau,\delta>0$, 
   the convergence rate of $\hat{f}_n$ satisfies 
   $$
   \norm{\hat{f}_n-f^*}_{L^2(\mathbb{P}_{X})} \leq  \sqrt{\frac{M \tilde{d}  (\tau+\delta)  }{n}}  
   $$
   with probability at least $1-2e^{-\tau}- e^{-\delta}$.
\end{mythm}

The above convergence rate in RKHSs of this type is independent of the dimensions of $X$ and $Z$.
Instead, it depends on the \emph{effective dimension} $\tilde{d}$. 
If $\tilde{d}$ is much smaller than $d$, 
it will be advantageous to regress in RKHSs.
The effective dimension represents the number of directions over the image of canonical feature maps on the data.
The dimension of $\mathcal{U}$ could be less than that of $\mathcal{H}$. 
\cref{prop: risk} shows that it suffices to consider the space $\mathcal{U}$ whose dimension is less than $\mathcal{H}$
as $u^*$ takes the form $u^*(y,z) = \mathbb{E}_{X|z}[f(X)]-y. $ 
If the dimensions of $\mathcal{H}$ and $\mathcal{U}$ are different, 
then the constant $\tilde{d}$ should be $\max\{ dim(\mathcal{H}), dim(\mathcal{U})  \}$.

Finally, we show that the dual method is optimal up to logarithmic factors. 
The following theorem is proved using Tsybakov's method and the tool of metric entropy. 

\begin{mythm}
   \label{thm: lower bound of finite dimensional spaces}
   Consider data $(X_i,Y_i)_{i=1}^n$ following the relationship $Y_i=f(X_i)+E_i$, 
   where $E_i$ is a Gaussian or truncated Gaussian noise,
   and $f$ is in a $\tilde{d}$-dimensional function space $\mathcal{H}$.
   For any estimation algorithm $\pi$, 
   there exists a function $f\in\mathcal{H}$ such that 
   $$
   \normX{f-\hat{f}^\pi_n} = \Omega\left(  \sqrt{ {\tilde{d} }/{n}}  \right)
   $$
   for the estimator $\hat{f}^\pi_n$ obtained by $\pi$ from the data.
\end{mythm}

\begin{myrem}
   The tool of metric entropy also plays a key role in describing infinite-dimensional spaces.
   The properties of infinite-dimensional spaces are more complex compared with finite-dimensional cases,
   so we discuss them in \cref{sec: concentration inequalities}.
\end{myrem}

   Note that in this paper, we assume the availability of instrumental variables.
   If the instrumental variable $Z$ is not accessible, 
   a constant lower bound can be reached in confounded settings. 
   It is because a bias will occur even for the best-$L^2(\mathbb{P}_X)$ predictor $\mean{Y|X}$.
   This \emph{non-identifiable} result necessitates further assumptions to deal with hidden confounders \cite{causaloverview}.

\subsection{The Design of Bandit Algorithms}

\begin{algorithm}[h]
	\renewcommand{\algorithmicrequire}{\textbf{Input:}}
	\renewcommand{\algorithmicensure}{\textbf{Output:}}
	\caption{Dual Instrumental Variable Regression with an Epoch Learning Strategy}
	\label{alg: DIV-BLS}
	\begin{algorithmic}[1]
      \Require epoch schedule $0=\tau_0<\tau_1<\tau_2<\cdots$, confidence parameter $\delta$, kernel functions $k,l$, tuning parameters $\eta,\eta_1,\eta_2$
      \State Determine $\tilde{d} $ from kernels $k$ and $l$
      \For{ epoch $m = 1,2,\cdots,$ }
         \State Collect (only) the data in epoch $m-1$, e.g., 
            $$
             \mathcal{D}_{m-1} = \{ (a_{\tau_{m-2}+1  }, c_{\tau_{m-2}+1  },y_{\tau_{m-2}+1  }, z_{\tau_{m-2}+1  }   ),  
             \cdots,  
             (a_{\tau_{m-1}  }, c_{\tau_{m-1}  },y_{\tau_{m-1}  }, z_{\tau_{m-1}  }   )   \}
             $$
         \State Let regularization parameters $\lambda_i = \eta_i\sqrt{ \tilde{d}/|\mathcal{D}_{m-1}|  }$ for $i=1,2$
         \State Implement dualIV with input $\lambda_1,\lambda_2$, $k,l$ and $\mathcal{D}_{m-1}$, 
                and then obtain $\hat{f}_m$ (for epoch $1$, $\hat{f}_1=0$)
         \State Compute $\gamma_m =  \sqrt{  \frac{\eta K (\tau_{m-1} -\tau_{m-2})  }{\tilde{d}  \log (2m^2/\delta)  }     }    $ (for the first epoch, $\gamma_1 = 1$)
            \For{ round $t = \tau_{m-1}+1,\cdots, \tau_{m}$ }
            \State Observe the context $c_t$ and the instrumental variable $z_t$
            \State  Compute $\hat{f}_m(c_t,a)$ for each action $a\in\mathcal{A}$ and the following probabilities
            $$
            p_t(a) = \left\{
            \begin{array}{lcl}
                \frac{1}{K+\gamma_m (\hat{f}_m (c_t, \hat{a}_t )- \hat{f}_m (c_t, a ) ) },  \text{ for all } a\neq \hat{a}_t \\
                1- \sum_{a\neq \hat{a}_t }  p_t(a), \text{ for } a = \hat{a}_t .
            \end{array}
            \right.
            $$
            where $\hat{a}_t =\max_{a\in\mathcal{A}} \hat{f}_m(c_t,a)   $.
            \State Sample $a_t \sim p_t(\cdot)$ and take the action $a_t$
            \State Observe a reward $y_t$
            \EndFor
         \EndFor
   \end{algorithmic}  
\end{algorithm}

A straightforward idea is to pick the arm 
$$
a_t =\argmax_{a\in\mathcal{A}} \hat{f}(c_t,a_t)
$$
during each round. 
However, the generated sequence is not i.i.d.\ because the action series definitely depends on the historical information. 
We apply the epoch learning strategy in (e.g., \cite{fasterCB,BeyondLS}) to address this challenge. 
This technique can avoid the complex construction of confidence bound. 
Instead of feeding all the previous data into dualIV,
we only feed the data in the previous epoch. 
This epoch strategy is partially due to technical reasons (\cref{thm: oracle inequality for finite rank} requires i.i.d.\ data), 
and we want to avoid a more complicated construction of martingales. 
Motivated by greedy algorithms, we design an action sampling policy based on the inverse gap weighting technique.
The sampling policy keeps fixed during an epoch though changes over epochs. 
Hence, we can obtain an i.i.d.\ sequence within each epoch, 
and elegantly balance exploration and exploitation as we move along the epochs.
We further prove that the regret of this algorithm is rate-optimal for properly selected tuning parameters.

As a consequence of the epoch learning strategy,
\cref{alg: DIV-BLS} must run in gradually increasing epochs, e.g., $\tau_m =2^m$ or $\tau_m =[ 2T^{1-2^{-m}} ]$.
Moreover, the \cref{alg: DIV-BLS} is computationally efficient 
because our algorithm only calls the subroutine dualIV at the beginning of each epoch.
The computational cost of the kernel inversion is the bottleneck \cite{kernelisedContextualBandits,EfficientKernelUCB},
so our algorithm shows advantages in computation efficiency compared with previous works.
If the epoch schedule $\tau_m =2^m$, then the number of calls of dualIV in \cref{alg: DIV-BLS} is $\mathcal{O}(\log T)$.
If the epoch schedule $\tau_m =\lfloor 2T^{1-2^{-m}} \rfloor$, 
then the number of calls of dualIV in \cref{alg: DIV-BLS} is $\mathcal{O}(\log\log T)$.

The regularization parameters should be properly selected, 
which depends on the data size $n$ and the RKHSs. 
If $\mathcal{H}$ and $\mathcal{U}$ are finite-dimensional, 
then optimal regularization parameters $\lambda_1 \simeq \sqrt{  \frac{\dim(\mathcal{U}) }{ n }  }$ and $\lambda_2 \simeq \sqrt{  \frac{\dim(\mathcal{H}) }{ n }  }$.
If $\mathcal{H}$ and $\mathcal{U}$ are infinite-dimensional, 
the choice of $\lambda_1$ and $\lambda_2$ relies on the decay rate of eigenvalues of RKHSs, 
as discussed in \cref{subsec: RKHS}. 
The tuning parameters $\eta_i$ for $i=1,2$ will implicitly influence the choice of $\eta$, 
and the tuning parameter $\eta$ is used to match the constant in \cref{thm: oracle inequality for finite rank}
as its value is difficult to deduce in theory.
The requirement ``match'' means $\eta<\frac{1}{M}$, where $M$ is the constant in \cref{thm: oracle inequality for finite rank}.
When $M$ is large, i.e., not confident in estimation, the tuning parameter $\eta$ should be small. 
For small $\eta$, the sampling policy $p_t(\cdot)$ will get close to uniform distribution, 
and the algorithm will become difficult to distinguish the optimal action in $\mathcal{A}$.
As the proof of \cref{thm: regret upper bound} shows, 
the leading constant of the first term in regret upper bound is proportional to $\eta^{-\frac{1}{2}}$.


\section{Regret Analysis}
\label{sec: regret analysis}

\paragraph{Regret Upper Bound.}
Based on the epoch learning strategy and \cref{thm: oracle inequality for finite rank}, 
the following event 
$$
\mathcal{E}_m=\Big\{ 
 \meanX{ (\hat{f}_m(x_t) - {f}^*(x_t) )^2 } \leq \frac{K}{4\gamma_m^2},\tau_{m-2}+1  \leq t \leq \tau_{m-1} |\mathcal{F}_{t-1}\Big\}
$$
happens with probability at least $1- \frac{\delta}{2m^2}$.
Therefore, by the union bound, the event $\bigcap_{m\geq 2}  \mathcal{E}_m  $ holds with probability at least $1-\delta/2$.
We can obtain the following expected regret upper bound on the event $\bigcap_{m\geq 2}  \mathcal{E}_m$.

\begin{mythm}
    \label{thm: regret upper bound}
    Suppose that \cref{asp: realizability,asp: invertibility,asp: continuity} hold in kernelized contextual bandit settings.
    Moreover, the epoch schedule is set to be $\tau_m =  2^m$, 
    and the tuning parameters are properly selected to match the constant in \cref{thm: oracle inequality for finite rank}.
    Then, the expected regret $Reg(T)$ of \cref{alg: DIV-BLS} is at most
    $$
     \mathcal{O}(2\sqrt{KT \tilde{d} \log (2\log^2(T) /\delta)} +  \sqrt{8T \log(2/\delta)} )
    $$
    with probability at least $1-\delta$.
\end{mythm}

If we set $\delta = \tilde{d}/T$, then the expected regret can be further reduced to 
$\mathcal{O}(\sqrt{K\tilde{d} T \log ( 2T\log^2(T) /\tilde{d})  })$ 
via the property of conditional expectation.
The regret of this order is rate-optimal if we ignore the $\mathcal{O}(\log\log T)$ term.
For the epoch schedule $\tau_m = \lfloor 2T^{1-2^{-m}} \rfloor$, 
the proof of upper bound is similar so we omit it here.

\paragraph{Regret Lower Bound.}
Now we consider the regret lower bound. 
We first consider general RKHSs. 
The key idea is to utilize the hidden linearity of RKHS. 
We convert the kernelized contextual bandit to a linear bandit by the canonical feature map $\phi$. 
Since $\mathcal{H}$ is $\tilde{d}$-dimensional, 
the lower bound of linear bandit in \cite{minimaxOptimalLin,ReductionLowerBound} can be applied.

\begin{mythm}
    \label{thm: regret lower bound}
Assume that $\mathcal{H}$ is $\tilde{d}$-dimensional. 
Moreover, $K\leq 2^{\tilde{d}/2}$ and $T\geq \tilde{d}(\log K)^{1+\epsilon} $ for any small constant $\epsilon>0$.
For any algorithm $\pi$, 
there exists a bandit instance with a reward function $f\in\mathcal{H}$ such that 
$$
Reg(T)\geq \Omega \left( \sqrt{ \tilde{d} T \log K \log(T/\tilde{d})  } \right).
$$
\end{mythm}

Compared with the lower bound in \cref{thm: regret lower bound},
our upper bound has a $\sqrt{K}$ factor rather than $\sqrt{\log K}$. 
It is because $\mathcal{H}$ may be some good function families, such as the collection of linear functions. 
The structure of linearity is important in regret analysis and can reduce the $\sqrt{K}$ dependence to $\sqrt{\log K}$.
However, \citet{CBwithPredictableRewards} point out that the factor $\sqrt{K}$ is unavoidable for general function spaces, 
even with the realizability assumption.
To convince the readers, we consider the following function space
\begin{equation}
    \label{eq: multiple reward function space}
\mathcal{H}=\{ g_a(c)\in \mathcal{G}| \forall c \in \mathcal{C}, \forall a\in\mathcal{A}  \},
\end{equation}
where  $\mathcal{G}$ is a $\tilde{d}$-dimensional RKHS.
Reward functions of this form are widely used in literatures,
including contextual bandits \cite{EstimationCB}, offline learning \cite{partiallyobservable} and Markov decision problems \cite{RLMultipleReward}.
Bandits in multiple-reward-function settings have such form reward functions. 
We prove that the lower bound of bandit problems with the above reward functions is $\sqrt{\tilde{d}KT}$.

The core idea is to construct a $[(\tilde{d}-1)(K-1)+1]$-armed bandit problem
whose regret is a lower bound of kernelized contextual bandits.
Then we can apply the lower bound of MAB problems to prove our result. 
We show that a problem under the multiple-parameter setting 
with dimension $\tilde{d}$ is equivalent to that under the single parameter setting with dimension $K\tilde{d}$ \citep{smoothedanalysis}.

\begin{mythm}
    \label{thm: a special case of regret lower bound}
    Suppose $K>1$ and $\tilde{d}>1$. 
    Consider the function space $\mathcal{H}$ of the form \eqref{eq: multiple reward function space}.
    Then for any algorithm $\pi$, 
    there exists a reward function $f \in \mathcal{H} $ such that 
    $$
    Reg(T)\geq \Omega\left(  \sqrt{\tilde{d}KT}  \right).
    $$
\end{mythm}

Moreover, we have the following high-probability lower bound. 
This bound also implies the near-optimality of our algorithm if we ignore $\log\log T$ in the regret upper bound.
\begin{mythm}
    \label{thm: a special case of high-probability regret lower bound}
Suppose $K>1$ and $\tilde{d}>1$. 
Consider the function space $\mathcal{H}$ of the form \eqref{eq: multiple reward function space}.
Then for any algorithm $\pi$ and $\delta \in (0,1)$ satisfying 
$$
T\delta \leq \sqrt{ T  K\tilde{d} \log \Big( \frac{1}{4\delta} \Big)  },
$$
there exists a reward function $f \in \mathcal{H} $ such that 
$$
Reg(T)\geq    \frac{1}{4} \min  \Big\{  T,  \sqrt{\tilde{d}KT  \log \Big(\frac{1}{4\delta} \Big) }  \Big\}  
$$
with probability at least $1-\delta$.
\end{mythm}


\section{Numerical Experiments}
\label{sec: numerical experiments}

In this section, we illustrate the regret of our bandit \cref{alg: DIV-BLS} in numerical experiments.
We modify the example in \cite{dualIV} from linear relationship to polynomial relationship, so the following relationships generate our data points:
$$
y= (\alpha^\top x +1 )^3 + e, c = \rho z + (1-\rho) e,
$$
where $e\sim\mathcal{N}(0,0.1)$ and $z\sim Uniform[0,1]^2$.
We set the number of arms $K=4$,
and four actions are sampled i.i.d. from $Uniform[-2,2]^2$.
For simplicity, the action set keeps fixed during the execution of our algorithm.
The unknown parameter $\alpha$ is uniformly chosen in $[0,1]^4$ at the initialization of bandit instances.
A small $\rho$ indicates a strong correlation between noise and contexts.
We let it vary in $0.01,0.1,0.25,0.5,0.7,0.95$ to demonstrate the performance of our algorithms in different confounded settings.

For the parameters in \cref{alg: DIV-BLS}, 
we select the epoch schedule $\tau_m=2^m$, 
the tuning parameters $\eta=200\rho^2,\eta_1=1,\eta_2=1$, 
and the confidence parameter $\delta=0.1$.
As we proved in \cref{thm: regret upper bound}, 
the reciprocal of $\eta$'s square root reflects the errors of our estimators. 
Hence, $\eta$ should be selected large when $\rho $ is large.
The kernel functions for two spaces are both polynomials: 
$
(w^\top w'+1)^3,
$
where $w$ (or $w'$) represents variables $x$ and $(y,z)$ for $k$ and $l$, respectively.
We run \cref{alg: DIV-BLS} on this bandit instance with a time horizon $T=2^{10}$,
and we repeat this process for $20$ times to approximate the expected regrets. 
The corresponding numerical result is illustrated in \cref{fig: regret curve}.
We then explore the numerical performance with an action number $K=10$. 
We let $\rho$ vary in $0.01,0.1,0.25,0.3,0.5,0.6,0.95$.
The final result is illustrated in \cref{fig: regret curve for K=10}.

\def \green {ForestGreen}
\def \blue {RoyalBlue}
\def \orange {orange}
\def \marksquare {square}
\def \marko {o}
\def \marktri {triangle}

\begin{figure}[hbtp]
    \centering
\begin{tikzpicture}
    \begin{axis}[
        height = 0.45\textwidth,
        width = 0.88\textwidth,
        xlabel = time horizon $ T $,   
        xmin = 0,
        ymin = 0,
        scaled ticks=false,
        ylabel = $Reg(T)$,
        scaled ticks=false,
        xtick pos = left,
        ytick pos = left,
        x tick label style={/pgf/number format/.cd,std,precision=0,
                            /pgf/number format/fixed,     
                            /pgf/number format/fixed zerofill,
                             },
        legend style={at={(0.03,0.7)}, anchor=west, nodes={scale=1, transform shape}}
    ]
    
    \addplot [dashed, line width = 1pt ] table [x index=0,y index=1, col sep = comma] {data/sample_20_10_0.01.csv};
    \addplot [color=black, line width = 1pt ] table [x index=0,y index=1, col sep = comma] {data/sample_20_11_0.1.csv};
    \addplot [color=\blue, line width = 1pt ] table [x index=0,y index=1, col sep = comma] {data/sample_20_11_0.25.csv};
    \addplot [color=purple, line width = 1pt ] table [x index=0,y index=1, col sep = comma] {data/sample_20_11_0.5.csv};
    \addplot [color=\orange, line width = 1pt ] table [x index=0,y index=1, col sep = comma] {data/sample_20_11_0.7.csv};
    \addplot [color=\green, line width = 1pt ] table [x index=0,y index=1, col sep = comma] {data/sample_20_11_0.95.csv};
 
    \legend{
        $\rho = 0.01$,
        $\rho = 0.1$,
        $\rho = 0.25$,
        $\rho = 0.5$,
        $\rho = 0.7$,
        $\rho = 0.95$,
    }
    \end{axis}
  
\end{tikzpicture}
    \caption{The regret curves for $K=4$.}

    \label{fig: regret curve}
\end{figure}
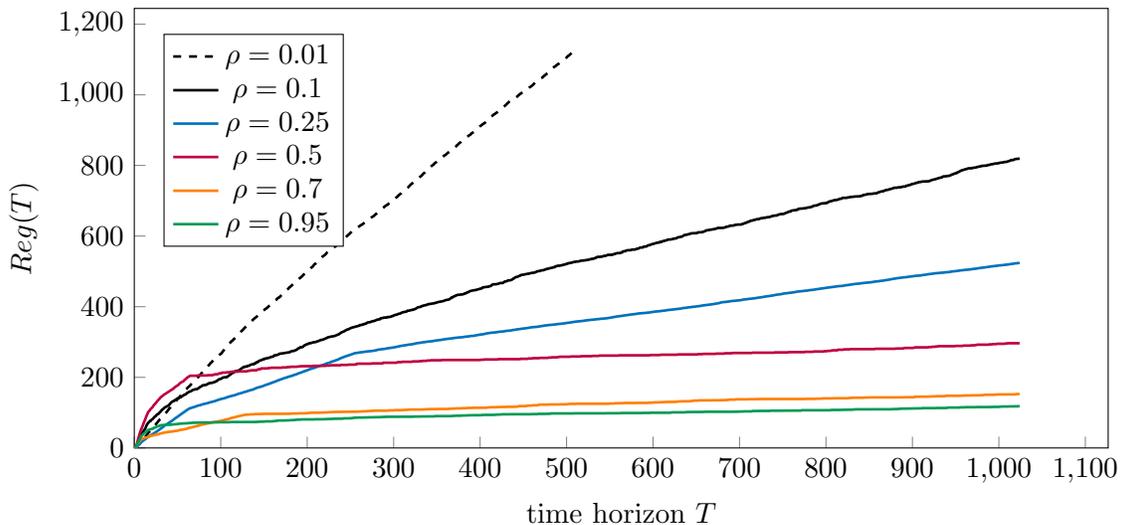

\begin{figure}[hbtp]
    \centering
\begin{tikzpicture}
    \begin{axis}[
        height = 0.45\textwidth,
        width = 0.88\textwidth,
        xlabel = time horizon $ T $,   
        xmin = 0,
        ymin = 0,
        scaled ticks=false,
        ylabel = $Reg(T)$,
        scaled ticks=false,
        xtick pos = left,
        ytick pos = left,
        x tick label style={/pgf/number format/.cd,std,precision=0,
                            /pgf/number format/fixed,     
                            /pgf/number format/fixed zerofill,
                             },
        legend style={at={(0.78,0.7)}, anchor=west, nodes={scale=1, transform shape}}
    ]
    
    \addplot [dashed, line width = 1pt ] table [x index=0,y index=1, col sep = comma] {data/sample_20_9_0.01_10.csv};
    \addplot [color=black, line width = 1pt ] table [x index=0,y index=1, col sep = comma] {data/sample_20_11_0.1_10.csv};
    \addplot [color=\blue, line width = 1pt ] table [x index=0,y index=1, col sep = comma] {data/sample_20_11_0.25_10.csv};
    \addplot [color=red, line width = 1pt ] table [x index=0,y index=1, col sep = comma] {data/sample_20_11_0.3_10.csv};
    \addplot [color=purple, line width = 1pt ] table [x index=0,y index=1, col sep = comma] {data/sample_20_11_0.5_10.csv};
    \addplot [color=yellow, line width = 1pt ] table [x index=0,y index=1, col sep = comma] {data/sample_20_11_0.6_10.csv};
    \addplot [color=\green, line width = 1pt ] table [x index=0,y index=1, col sep = comma] {data/sample_20_11_0.95_10.csv};
 
    \legend{
        $\rho = 0.01$,
        $\rho = 0.1$,
        $\rho = 0.25$,
        $\rho = 0.3$,
        $\rho = 0.5$,
        $\rho = 0.6$,
        $\rho = 0.95$,
    }
    \end{axis}
  
\end{tikzpicture}
    \caption{The regret curves for $K=10$.}

    \label{fig: regret curve for K=10}
\end{figure}
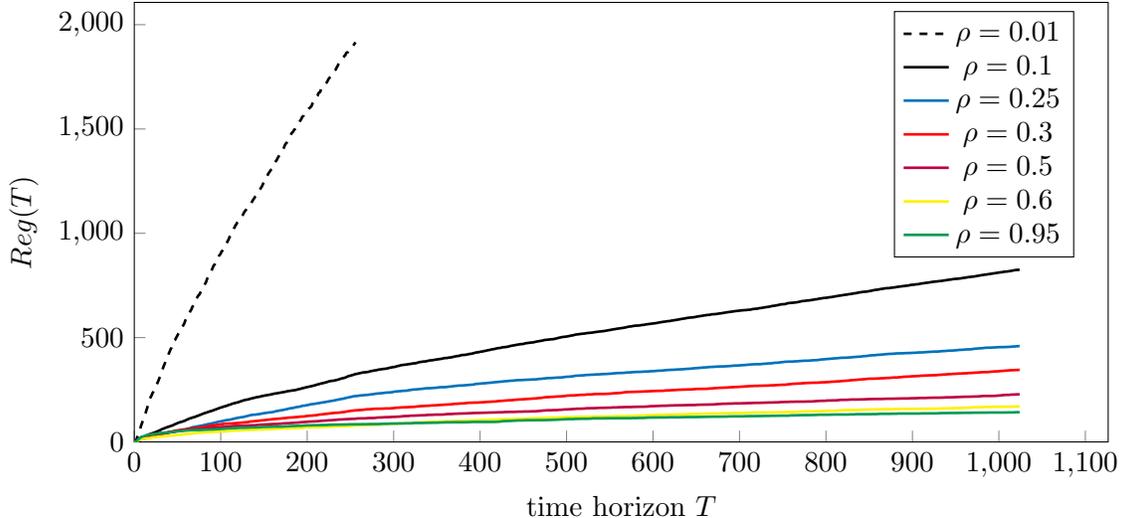

The tendency of the regret curves in \cref{fig: regret curve} and \cref{fig: regret curve for K=10}
show the efficacy of our proposed \cref{alg: DIV-BLS}. 
The dashed curve serves as a benchmark in \cref{fig: regret curve} 
because $\rho =0.01$ is close to the case where instrumental variables are unavailable.
A weak instrumental variable and a strong confounder can induce almost linear regret. 
However, \cref{fig: regret curve} and \cref{fig: regret curve for K=10} illustrate that even a weak instrumental variable can reduce the regret from linear to sub-linear.
When $\rho$ becomes large, the slope of the curves gradually flattens out, 
suggesting that the dualIV estimator indeed makes good predictions on the rewards of each action.
These numerical results clearly show the advantages of obtaining causal knowledge.
Our algorithm can fully utilize the information brought by instrumental variables to achieve optimal regrets
as it is tailored to such confounded settings.

\section{Future Directions}
First, we assume that an instrumental variable is accessible to the learner, 
because finding an instrumental variable is not the focus of our work.
It may be interesting to propose a practical principle of finding instrumental in specific problems, such as \cite{RecommendationInstrument}.
Second, we consider the IV regression in a realizable setting. 
It will be meaningful to solve the problem in an agnostic setting.
Then, the epoch learning strategy avoids constructing confidence bound for estimators, 
but this strategy is sample-inefficient. 
A future direction is to propose the sample-efficient algorithm which utilizes all the data obtained from previous rounds.
One possible idea is to apply a martingale technique as \citet{improvedUCB} do, 
in order to update a confidence upper bound for estimators adaptively. 
This technique may improve the regret upper bound from $\sqrt{K}$-dependence to $\sqrt{\log K}$-dependence in kernelized bandits.
Intuitively, the linearity (or the controlled non-linearity) in RKHSs plays a key role in regret analysis. 
Finally, the problem about the lower bound for general kernelized bandit remains unsolved, 
though the lower bound in RKHSs of certain metric entropy has been proved by \citet{CBwithRegressionOracles}.
A possible method is to apply the concept of effective dimension proposed in the literatures \cite{EffectiveDimension,kernelisedContextualBandits,EfficientKernelUCB}
to deduce lower bounds in a unified way.

\newpage

\bibliographystyle{plainnat}
\bibliography{ref}

\newpage
\appendix

\section{Dual Method}
\label{sec: dual method}

Given the dataset $\{x_i,y_i,z_i\}_{i=1}^n$,
define $\Phi = [ \phi(x_1),\cdots,\phi(x_n)  ]$, $\Upsilon  = [ \varphi(y_1,z_1),\cdots,  \varphi(y_n,z_n)]$ and $\textbf{y}=[y_1,\cdots,y_n]^T$.
Then, the quantities $r, \covYZ, \covXYZ$ can be estimated by 
$$
\hat{r} = \frac{1}{n} \sum_{i=1}^n y_i \varphi(y_i,z_i) = \frac{1}{n} \Upsilon \textbf{y}, 
$$
$$
\hat{\mathfrak{C}}_{YZ} = \frac{1}{n} \sum_{i=1}^n \varphi(y_i,z_i)\otimes \varphi(y_i,z_i) = \frac{1}{n}\Upsilon \Upsilon^T,
$$
and 
$$
\hat{\mathfrak{C}}_{XYZ} = \frac{1}{n} \sum_{i=1}^n \phi(x_i) \otimes  \varphi(y_i,z_i) = \frac{1}{n} \Phi \Upsilon^T.
$$
Hence, the estimator is 
$$
\hat{f} = \left( \hat{\mathfrak{C}}_{XYZ} (\hat{\mathfrak{C}}_{YZ}+\lambda_1\mathcal{I})^{-1} \hat{\mathfrak{C}}_{YZX} +\lambda_2\mathcal{I}  \right)^{-1} \hat{\mathfrak{C}}_{XYZ} (\hat{\mathfrak{C}}_{YZ}+\lambda_1\mathcal{I})^{-1}\hat{r}.
$$ 
   
The dual instrumental variable regression can be simplified by the following algorithm \cite{dualIV}:

\begin{algorithm}
	\renewcommand{\algorithmicrequire}{\textbf{Input:}}
	\renewcommand{\algorithmicensure}{\textbf{Output:}}
	\caption{dualIV}
	\label{alg: dualIV}
	\begin{algorithmic}[1]
      \Require Dataset $(x_i,y_i,z_i)_{i=1}^n$, kernel functions $k,l$, parameters $\lambda_1,\lambda_2$
		\State Compute kernel matrices $\textbf{K}_{ij}=k(x_i,x_j)$, $\textbf{L}_{ij}=k((y_i,z_i),(y_j,z_j)) $
      \State  $\textbf{M} \leftarrow \textbf{K} (\textbf{L}+ n \lambda_1 \textbf{I} )^{-1}\textbf{L}    $
      \State  $\theta \leftarrow  (\textbf{M}   \textbf{K} + n \lambda_2 \textbf{K} )^{-1}\textbf{M}\textbf{y}    $
		\Ensure  $f(x) = \Phi  \theta $
   \end{algorithmic}  
\end{algorithm}

We then establish some properties of $\Psi(f,u)$ in dual method. 

\begin{myprop}
   \label{prop: risk}
   Assume $f^*$ is the solution to the minimax problem. 
\begin{enumerate} 
   \item For any fixed $f$, the maximizer of inner optimization problem is $u^*(y,z)=\mathbb{E}_{X|z}[f(X)]-y$.
   \item For any fixed $f$, $R(f)= \max_{u} \Psi(f,u)$.
   \item $\frac{1}{2}\norm{f-f^*}^2_{L^2(\mathbb{P}_X)} = R(f)-R(f^*)=  \max_u\Psi(f,u)-\max_u\Psi(f^*,u)$. 
   \item $\frac{1}{2}\norm{u-u^*}^2_{L^2(\mathbb{P}_{YZ})} = \Psi(f,u^*)-\Psi(f,u)$ for any given $f$ and its corresponding maximizer $u^*$.
\end{enumerate}
\end{myprop}

\begin{myproof}
\begin{enumerate}
   \item Given $f$, $\Psi(f,u)$ is an unconstrained quadratic program with respect to $u$. Since
   $$
   \Psi(f,u) = -\frac{1}{2}\mathbb{E}_{XYZ}[ (u(Y,Z) - f(X) +Y)^2 ] + \mathbb{E}_{XYZ}[ ( f(X) -Y)^2 ], 
   $$
   the maximizer $u^*(Y,Z)$ takes the form $\mathbb{E}_{X|YZ}[f(X)-Y]$. Hence, 
   $$
   u^*(y,z)=\mathbb{E}_{X|z}[f(X)]-y. 
   $$

   \item 
   Plugging in $u^*$ in $\Psi(f,u)$, we have 
   $$
   \begin{array}{rcl}
   \Psi(f,u^*) &=& \mathbb{E}_{XYZ}[(f(X)-Y)u^*(Y,Z)]-\frac{1}{2} \mathbb{E}_{YZ}[u^*(Y,Z)^2] \\
   &=& \mathbb{E}_{XYZ}[(f(X)-Y) (\mathbb{E}_{X|Z}[f(X)]-Y)   ]-\frac{1}{2} \mathbb{E}_{YZ}[(\mathbb{E}_{X|Z}[f(X)]-Y)^2] \\
   &=& \mathbb{E}_{YZ}[ (\mathbb{E}_{X|Z}[f(X)]-Y)(\mathbb{E}_{X|Z}[f(X)]-Y)   ]-\frac{1}{2} \mathbb{E}_{YZ}[(\mathbb{E}_{X|Z}[f(X)]-Y)^2] \\
   &=& \frac{1}{2} \mathbb{E}_{YZ}[(\mathbb{E}_{X|Z}[f(X)]-Y)^2] \\   
   &=& R(f)          
   \end{array}
   $$

   \item From 2, 
   $$
   \begin{array}{rcl}
      & & 2[R(f) - R(f^*)] \\
      &=& 2[\Psi(f,u^*) - \Psi(f^*,u^*)] \\ 
      &=&  \mathbb{E}_{YZ}[(\mathbb{E}_{X|Z}[f(X)]-Y)^2] -  \mathbb{E}_{YZ}[(\mathbb{E}_{X|Z}[f^*(X)]-Y)^2]   \\ 
      &=&  \mathbb{E}_{YZ}[(\mathbb{E}_{X|Z}[f(X)])^2 - 2Y\mathbb{E}_{X|Z}[f(X)] - (\mathbb{E}_{X|Z}[f^*(X)])^2 + 2Y\mathbb{E}_{X|Z}[f^*(X)]   ]   \\ 
      &=& \mathbb{E}_{X}[f(X)]^2 -  \mathbb{E}_{X}[f^*(X)]^2  - 2\mathbb{E}_{YZ}[ Y\mathbb{E}_{X|Z}[f(X)] - Y\mathbb{E}_{X|Z}[f^*(X)] ]   \\ 
      &=& \mathbb{E}_{X}[f(X)]^2 -  \mathbb{E}_{X}[f^*(X)]^2  - 2 \mathbb{E}_{XYZ}[Yf(X)|Z] +2\mathbb{E}_{XYZ}[Yf^*(X)|Z]    \\ 
      &=& \mathbb{E}_{X}[f(X)]^2 -  \mathbb{E}_{X}[f^*(X)]^2  - 2 \mathbb{E}_{X}[f(X)f^*(X) ] +2 \mathbb{E}_{X}[f^*(X)]^2    \\ 
      &=& \mathbb{E}_{X}[f(X)-f^*(X)]^2    \\ 
      &=& \norm{f-f^*}^2_{L^2(\mathbb{P}_X)}
   \end{array}
   $$

   \item From 1, 
   $$
   \begin{array}{rcl}
      & &   2[\Psi(f,u^*) - \Psi(f,u)] \\ 
      &=&  \mathbb{E}_{XYZ}[(u^*(Y,Z)-f(X)+Y )^2] -  \mathbb{E}_{XYZ}[(u(Y,Z)-f(X)+Y )^2]  \\ 
      &=&  \mathbb{E}_{XYZ}[u^*(Y,Z)^2- u(Y,Z)^2 +2(Y-f(X)) (u^*(Y,Z)-u(Y,Z)) ] \\
      &=& \mathbb{E}_{YZ}[u^*(Y,Z)^2- u(Y,Z)^2 +2 \mathbb{E}_{X|Z} (Y-f(X)) (u^*(Y,Z)-u(Y,Z)) |Z ] \\
      &=& \mathbb{E}_{YZ}[u^*(Y,Z)^2- u(Y,Z)^2 +2 u^*(Y,Z) (u^*(Y,Z)-u(Y,Z)) |Z ] \\
      &=& \mathbb{E}_{YZ}[u^*(Y,Z)^2 + u(Y,Z)^2 -2 u^*(Y,Z) u(Y,Z) |Z ] \\
      &=& \norm{u-u^*}^2_{L^2(\mathbb{P}_{YZ})}
   \end{array}
   $$

\end{enumerate}
\end{myproof}

\section{Reproducing Kernel Hilbert Spaces}
\label{subsec: RKHS}

We provide some background knowledge on reproducing kernel Hilbert spaces (RKHSs) in this section. 
Given a subset $\mathcal{X}\subset \mathbb{R}^d$ and a probability measure $\mathbb{P}_X$ on $\mathcal{X}$, 
we consider a Hilbert space $\mathcal{H}\subset L_2(\mathbb{P}_X) $ associated with inner product $\innerH{\cdot}{\cdot}$. 
\begin{mydef}
A Hilbert space $\mathcal{H}$ of functions defined on a nonempty set $\mathcal{X}$ is said to be a reproducing kernel Hilbert space 
if the Dirac evaluation functional $\delta_x$ is continuous $\forall x \in \mathcal{X}$.
\end{mydef}
\begin{mydef}
Let $\mathcal{H}$ be a Hilbert space of $\mathbb{R}$-valued functions defined on a nonempty set $\mathcal{X}$. 
A function $k:\mathcal{X}\times \mathcal{X} \to \mathbb{R}$ is called a reproducing kernel of $\mathcal{H}$ if it satisfies 
\begin{enumerate}
\item $\forall x\in \mathcal{X}, k(\cdot,x)\in \mathcal{H}$
\item $\forall x\in \mathcal{X}, \innerH{f}{k(\cdot,x)}=f(x)$(the reproducing property).
\end{enumerate}
\end{mydef}
Riesz's representation theorem implies the reproducing kernel $k$ of a RKHS $\mathcal{H}$ uniquely exists.
Such kernel must be positive semidefinite. 
If $k$ is continuous, 
Mercer's theorem guarantees that the kernel has an eigen-expansion of the form 
$$
k(x,x') = \sum_{i=1}^\infty \mu_i b_i(x)b_i(x'), 
$$
where $\mu_1\geq\mu_2\geq\cdots\geq 0$ is a non-negative decreasing sequence of eigenvalues (of the integral operator defined by the kernel), 
and $\{b_i\}_{i=1}^\infty$ are the associated eigenfunctions, taken to be orthonormal.  
Since $\{b_i\}_{i=1}^\infty$ form an orthonormal basis, 
any function $f\in\mathcal{H}$ has an expansion $f(x) = \sum_{i=1}^\infty \theta_i b_i(x)$.

Then we investigate the decay rate of eigenvalue sequence $\{\mu_i\}_{i=1}^\infty$, 
which plays a crucial role in our analysis. 
We first discuss the cases with finitely nonzero $m$ eigenvalues, 
meaning $\mathcal{H}$ is a $m$-dimensional space.  
We call such kernels are \emph{of finite rank}. 
Conversely, the kernels of the RKHSs with infinitely many nonzero eigenvalues are \emph{of infinite rank}. 
For the latter, we mainly consider the kernels with eigenvalues whose decay rate is at $\mu_i \simeq i^{-2\nu/d}$. 
The notation $\simeq$ means two quantities are of the same order up to a constant.

The covariance and cross-covariance operators on RKHSs are also important concepts for modern applications of RKHSs. 
In principle, they are generalizations of covariance and cross-covariance matrices in Euclidean
space to the infinite-dimensional elements in RKHSs.
The (uncentered) cross-covariance $\covYZX$ is the linear operator
which maps $\mathcal{H}$ to $\mathcal{U}$.
It has several equivalent definitions, 
and in this paper, we define the operator in terms of the tensor product $\mathbb{E}_{XYZ}[ \varphi(Y,Z) \otimes \phi(X)  ] $. 
Then for all $f\in\mathcal{H} $, 
$$
\covYZX f = \mathbb{E}_{XYZ}[  \innerH{\phi(X)}{f}\varphi(Y,Z)  ].
$$
$ \covYZX$ is the adjoint of $\covXYZ$. 
It can be shown using Hilbert-Schmidt theory that 
$$
\innerH{ \covXYZ u }{f} = \innerU{ \covYZX f }{u} = \meanXYZ{ f(X) u(Y,Z) }.
$$
This property indicates that the cross-covariance operator captures the covariance of two elements in RKHSs.
Similarly, we can define the covariance operator 
$$
\covYZ = \mathbb{E}_{YZ}[ \varphi(Y,Z) \otimes \varphi(Y,Z)  ].
$$
It is self-adjoint and maps from $\mathcal{U}$ to $\mathcal{U}$.

For more knowledge on reproducing kernel Hilbert space, 
we refer readers to the book \cite{RKHStheory} and the review \cite{Kernelmeanembedding}.

\subsection{Boundedness lemmas}

Under the assumptions \eqref{asp: invertibility} and \eqref{asp: continuity}, 
the norm of covariance operators and their inverse operators are bounded. 
We show this result in the following lemma. 

\begin{mylem}
    \label{lem: boundedness of covariance operators}
 Under the assumptions \eqref{asp: invertibility} and \eqref{asp: continuity}, 
 the operators $ \covXYZ,\covYZX,\covYZ$ defined on RKHSs are bounded. 
 Moreover, their inverse operators are also bounded.
\end{mylem}
\begin{myproof}
The linear operator $ \covXYZ$ maps $\mathcal{U}$ to $\mathcal{H}$. 
Let 
$$
\kappa=\max\left\{ \max_{x\in\mathcal{X}} k(x,x), \max_{(y,z)\in\mathcal{Y}\times\mathcal{Z} } l((y,z),(y,z))   \right\}, 
$$
because $k$ and $l$ are continuous functions on compact sets.
Let $u\in\mathcal{U}$, then 
$$
\begin{array}{rcl}
    \normX{ \covXYZ u }&=& \normX{  \meanXYZ{ \innerH{ \varphi(Y,Z) }{  u  } \phi(X)  } }\\ 
                       &\leq&  \kappa \normYZ{u}. \\
\end{array}
$$
Hence, $\covXYZ$ is a bounded linear operator. 
The boundedness of remaining linear operators can be proved similarly by their definitions. 
$\mathcal{H}$ and $\mathcal{U}$ are Banach spaces as they are RKHSs. 
Thanks to the Banach bounded inverse theorem, 
the inverse of linear operators mapping from one Banach space to another are bounded. 
Therefore, the inverse operators of $ \covXYZ,\covYZX,\covYZ$ are bounded.

\end{myproof}

Thanks to \cref{lem: boundedness of covariance operators}, 
the following concentration inequalities hold.
We apply the Hoeffding's inequality to finish the proof.
\begin{mylem}
   \label{lem: concentration of op}
Suppose that continuous $k$ and $l$ are bounded by $\kappa_1,\kappa_2$, i.e., 
$$
\norm{k}_\infty:=\sup_{x\in\mathcal{X}} k(x,x) \leq \kappa_1, \norm{l}_\infty:= \sup_{(y,z)\in\mathcal{Y}\times \mathcal{Z}}l ((y,z),(y,z)  ) \leq \kappa_2. 
$$
Then for any given continuous function $h\in \mathcal{H}$,
$$
\Prob{ \normYZ{   \hat{\mathfrak{C}}_{YZX}h - \covYZX h   }  \geq  \sqrt{\frac{B\kappa_1\kappa_2   \log (1/\delta)}{2n}}  } \leq \delta  ,
$$
where $B$ is the bound of $h$.
\end{mylem}

\begin{myproof}
Let $\xi_i = \innerH{h}{\phi(x_i)} \varphi (y_i,z_i) $, 
and the random function $\xi_i$ maps $\mathcal{Y}\times\mathcal{Z}$ to $\mathbb{R}$.
To avoid the ambiguity, we use the variable $w$ that takes the value in $\mathcal{Y}\times\mathcal{Z}$ instead of $(y,z)$.
Then 
$$
\begin{array}{rcl}
\mathbb{E}[ \xi_i  ] &=& \mathbb{E}[ \innerH{h}{\phi(x_i)} \varphi (y_i,z_i) ]  \\
&=& \mathbb{E}[  \varphi (y_i,z_i) \otimes \phi(x_i)  ] h  \\
&=& \mathbb{E}_{XYZ}[\varphi (Y,Z)  \otimes \phi(X)  ] h \\
&=& \covYZX h. \\
\end{array}
$$
Since $\xi_i(w)$ is bounded by $ B\kappa_1 \kappa_2$, then for any given $w\in\mathcal{Y}\times\mathcal{Z}$,  
$$
\Prob{ \bigg|  (\frac{1}{n} \sum_{i=1}^n\xi_i - \covYZX h )(w)   \bigg|  \geq \sqrt{\frac{B\kappa_1\kappa_2   \log (1/\delta)}{2n}} }       \leq \delta.
$$
by Hoeffding's inequality.
Clearly, $\frac{1}{n} \sum_{i=1}^n\xi_i - \covYZX h $ is a continuous function on $\mathcal{Y}\times\mathcal{Z}$. 
By the intermediate value theorem, there exists $w^* \in \mathcal{Y}\times\mathcal{Z}$ such that
$$
\normYZ{\frac{1}{n} \sum_{i=1}^n\xi_i - \covYZX h }   = \left| \bigg(\frac{1}{n} \sum_{i=1}^n\xi_i - \covYZX h \bigg)(w^*)   \right| 
$$
Let $w=w^*$ and the following holds
$$
\Prob{ \normYZ{   \hat{\mathfrak{C}}_{XYZ}h - \covYZX h   }  \geq  \sqrt{\frac{B\kappa_1\kappa_2   \log (1/\delta)}{2n}}  } \leq \delta  .
$$

\end{myproof}

\section{Concentration Inequalities}
\label{sec: concentration inequalities}

\subsection{Proof of the oracle inequality}

We defer the proof of \cref{thm: oracle inequality for finite rank} at the end of this subsection, 
because we need to prepare some useful lemmas to complete the proof.
\citet{optRatesRLSA,svm} and other scholars prove the optimal rates for regularized least squares regression. 
They consider a non-parametric least squares regression: 
$$
\mathcal{R}(f) = \int_{\mathcal{X}\times\mathcal{Y}} (y-f(x))^2 d \mathbb{P}(x,y). 
$$
Denote 
\begin{equation}
   \label{eq: quadratic loss}
\hat{\mathcal{R}}(f) = \frac{1}{n}\sum_{i=1}^n (y_i-f(x_i))^2
\end{equation}
as the empirical version of $\mathcal{R}(f)$.
Then the estimator is solved by 
\begin{equation}
   \label{eq: f solved in quadratic loss}
\hat{f} \in \argmin_{f\in\mathcal{H}} \lambda \norm{f}_\mathcal{H}^2 + \hat{\mathcal{R}}(f).
\end{equation}


Many papers, including \cite{BeyondLS}, only achieves suboptimal convergence rate. 
To achieve better rates, many novel techniques are applied in literatures. 
\citet{optRatesRLSR,svm} applies the trick of clipping. 
The idea, by nature, is to find a solution in a closed ball of the original spaces, 
which can simplify the process of solving regression problems in RKHSs.
Denote $\wideparen{f}$ as the clipped value of $f$ at $\pm B$, that is 
    $$
    \wideparen{f} = \left\{
    \begin{array}{lcl}
    -B, & &f<-B\\
    t,  & &f\in[-B,B]\\
    B,  & &f>B.\\
    \end{array}
    \right.
    $$
The concentration results will hold for a clipping function. 
However, this method is not applicable in our setting, 
because the relationship \eqref{eq: relationship between f and u} does not hold for clipped functions.
For finite-dimensional cases, \citet{optRatesRLSA} achieves the optimal rates up to logarithmic terms.


\begin{mythm}[A simplifed version in \cite{optRatesRLSA}]
   \label{thm: rate of OLS for finite rank kernel}
   Let the function $k$ associated with $\tilde{d}$-dimensional RKHS $\mathcal{H}$ is a bounded measurable kernel on $\mathcal{X}$.
   Moreover, assume that $\norm{f}_{\infty}$ are bounded by $B$.
   Then, there exists a constant $c_{B}$ which depends on $B$ and the bound of $Y$, 
   such that for all $\tau>0$ and $\lambda \simeq \sqrt{\tilde{d}\tau /n } $, 
   the learning method described by \eqref{eq: f solved in quadratic loss} satisfies 
   $$
   \mathcal{R}(\hat{f}) - \mathcal{R}({f}^*) \leq   \frac{ c_B \tilde{d} \tau  }{n}  
   $$
   with probability not less than $1-e^{-\tau}$.
\end{mythm}

Now we try to utilize above theorem to show the convergence rate of our method. 
One challenge is that the problem considered in \eqref{eq: f solved in quadratic loss} is convex, 
while the dual kernel method need to solve a saddle point problem. 
Hence, $\Psi(f,u)$ is not a loss so we cannot directly apply \cref{thm: rate of OLS for finite rank kernel}.
However, $\Psi(f,u^*)=R(f)$ is a quadratic loss for $f$, 
and for any given $f$, $-\Psi(f,u)$ can be considered as a quadratic loss for $u$ with a constant shift.

\begin{mythm}
   Let the $\tilde{d}$-dimensional RKHSs $\mathcal{H}$ and $\mathcal{U}$ associated with kernels $k$ and $l$ satisfying \cref{asp: realizability,asp: invertibility,asp: continuity}.
   Consider a dataset $(X_i,Y_i,Z_i)_{i=1}^n$ i.i.d.\ sampled according to \cref{fig: causal model with instrumental variables}, 
   and $\hat{f}_n$ is obtained from dualIV with regularization parameters $\lambda_i \simeq  \sqrt{ \frac{  \tilde{d} \tau   }{n} }  $ for $i=1,2$.
   Then, there exists a constant $M$ 
   which depends on the true structural function $f^*$ and spaces $\mathcal{H},\mathcal{U}$,
   such that for all $\tau,\delta>0$, 
   the convergence rate of $\hat{f}_n$ satisfies 
   $$
   \norm{\hat{f}_n-f^*}_{L^2(\mathbb{P}_{X})} \leq  \sqrt{\frac{M \tilde{d}  (\tau+\delta)  }{n}}  
   $$
   with probability at least $1-2e^{-\tau}- e^{-\delta}$.
\end{mythm}

\begin{myproof}
   We mainly apply \cref{thm: rate of OLS for finite rank kernel} to $\hat{u}$. 
   By the definition, 
   $(\hat{f},\hat{u})$ is solved from 
   $$
   \min_{f\in\mathcal{H}}\max_{u\in\mathcal{U}}\hat{\Psi}_\lambda(f,u).
   $$
   Let the risk $\mathcal{R}$ be $-\Psi(\hat{f},u)$ and its empirical version is exactly $\hat{\Psi}_\lambda(\hat{f},u)$ due to \cref{prop: risk}.
   Since $\lambda_1 \simeq \sqrt{  \frac{\tilde{d} }{n }} $, 
   \begin{equation}
      \label{eq: error for u}
   \begin{array}{rcl}
   2\normYZ{\hat{u}-\hat{u}^*}^2  &=&  \Psi(\hat{f},\hat{u}^*) - \Psi(\hat{f},\hat{u})  \\
   &\leq&   \frac{c_{B}\tilde{d}\tau}{n}
   \end{array}
   \end{equation}
   with probability at least $1-e^{-\tau}$. 
   The equality is a result of \cref{prop: risk}, 
   and the inequality is due to \cref{thm: rate of OLS for finite rank kernel}. 
   
   Then, we apply  \cref{thm: rate of OLS for finite rank kernel} to $R(f)$. 
   However, the empirical version is $\hat{R}(f) = \hat{\Psi}(\hat{f},\hat{u}^*)$. 
   Hence, the solution to 
   $$
   \min_{f\in\mathcal{H}} \hat{R}(f) + \frac{\lambda_2}{2} \normH{f}^2 
   $$
   is not $\hat{f}$. We denote the solution as $\hat{f}^*$. 
   Similarly, as $\lambda_2 \simeq \sqrt{  \frac{\tilde{d} \tau }{n }} $, then 
   \begin{equation}
      \label{eq: error for f}
      \begin{array}{rcl}
      2 \normX{\hat{f}^*-f^*}^2  &=& \Psi(\hat{f},\hat{u}^*) - \Psi(f^*,u^*) \\
      &\leq&    \frac{c_{B}\tilde{d}\tau}{n}
      \end{array}
   \end{equation}
   with probability at least $1-e^{-\tau}$.

   Now we need to establish the relationship between $\hat{f}^*$ and $\hat{f}$.
   We consider the empirical version of \cref{eq: relationship between f and u} with regularization terms: 
   \begin{equation}
      (\hat{\mathfrak{C}}_{YZ}+\lambda_1\mathcal{I}) \hat{u}^* = \hat{\mathfrak{C}}_{YZX}\hat{f}^* - \hat{r}
   \end{equation}
   and 
   \begin{equation}
      (\hat{\mathfrak{C}}_{YZ}+\lambda_1\mathcal{I}) \hat{u} = \hat{\mathfrak{C}}_{YZX} \hat{f} - \hat{r}.
   \end{equation}
   Combining above equalities, the error norm $\normX{\hat{f}^*- \hat{f}} $ can be bounded by $\normYZ{\hat{u}- \hat{u}^*}$. 
   
   Denote the bound of kernels $k$ and $l$ as $\kappa_1$ and $\kappa_2$, respectively. 
   Suppose that the function and its estimator are bounded by a universal constant $B$ almost surely.
   For any given $\delta>0$, define the following event
   $$
   \mathcal{E} = \left\{ \normYZ{  \covYZX (\hat{f}^*- \hat{f})  }  \leq   \normYZ{ \hat{\mathfrak{C}}_{YZX} (\hat{f}^*- \hat{f})   } + \sqrt{\frac{\tilde{d} B\kappa_1\kappa_2   \delta}{n}}    \right\} .
   $$
   Let $h=\hat{f}^*- \hat{f}$ in  \cref{lem: concentration of op},
   and $h$ is bounded by $2B$.
   As $\hat{f}^*$ and $ \hat{f}$ can be written as a linear combination of canonical feature maps,
   the continuity of kernels implies the continuity of $h$. 
   Thanks to \cref{lem: concentration of op}, 
   $\prob{\mathcal{E} }\leq e^{-\delta}$.
   Since we assume $\covYZ$ and $ \covXYZ \covYZ^{-1} \covYZX$ are invertible, 
   $\covXYZ \covYZX$ is also invertible.
   Then condition on the event $\mathcal{E}$, we have
   \begin{equation}
      \label{eq: error control}
   \begin{array}{rcl}
      &  &\normX{\hat{f}^*- \hat{f}}  \\
      &\leq& \norm{ (\covXYZ \covYZX)^{-1}\covXYZ  }_{op}  \normYZ{  \covYZX (\hat{f}^*- \hat{f})    } \\
      &\leq& \norm{ (\covXYZ \covYZX)^{-1}\covXYZ  }_{op}    \normYZ{ \hat{\mathfrak{C}}_{YZX} (\hat{f}^*- \hat{f})    } + \sqrt{\frac{B\kappa_1\kappa_2   \delta}{n}}\norm{  (\covXYZ \covYZX)^{-1} \covXYZ}_{op}  \\
      &\leq& \norm{ (\covXYZ \covYZX)^{-1}\covXYZ  }_{op}     \norm{ \hat{\mathfrak{C}}_{YZ} +\lambda_1\mathcal{I}}_{op} \normYZ{ \hat{u}- \hat{u}^*    }+ \sqrt{\frac{ \tilde{d}B\kappa_1\kappa_2 \delta }{n}}\norm{ (\covXYZ \covYZX)^{-1}\covXYZ }_{op}   \\
      &\leq&  (B^2 + \gamma_1)\norm{ (\covXYZ \covYZX)^{-1} }_{op}    \normYZ{ \hat{u}- \hat{u}^*    } + \sqrt{\frac{\tilde{d}B\kappa_1\kappa_2 \delta  }{n}}\norm{ (\covXYZ \covYZX)^{-1}\covXYZ }_{op} .
   \end{array}
   \end{equation}
The regularization parameter $\lambda_1 $ is bounded by a constant $\gamma_1$ as it converges to $0$.

   Applying the inequality $\sqrt{a+b}\leq \sqrt{a}+\sqrt{b}$ to \eqref{eq: error for f} and \eqref{eq: error for u}, 
   and combining the inequality \eqref{eq: error control}, the following holds: 
   \begin{equation}
      \begin{array}{rcl}
      & & \norm{\hat{f}-f^*}_{L^2(\mathbb{P}_{X})} \\
      &\leq& \normX{\hat{f}^*-\hat{f}} + \normYZ{\hat{f}^*-f^*} \\
      &\leq&  (B^2 + \lambda_1)\norm{ (\covXYZ \covYZX)^{-1}\covXYZ  }_{op}   \normYZ{ \hat{u}- \hat{u}^*    }  \\
      & \ & +\sqrt{\frac{\tilde{d}B\kappa_1\kappa_2 \delta  }{n}}\norm{ (\covXYZ \covYZX)^{-1}\covXYZ  }_{op}  + \normX{\hat{f}^*-f^*}   \\
      &\leq&   \sqrt{\frac{c \tilde{d} (\tau +\delta) }{n}}  
      \end{array}
   \end{equation}
   with probability at least $1-2e^{-\tau} - e^{-\delta}$.
   Here, the constant $M$ is chosen as 
   $$
   \max\{ \norm{ (\covXYZ \covYZX)^{-1}\covXYZ  }_{op}  \sqrt{c_B}( B^2 +\gamma_1  ), \sqrt{ {B\kappa_1\kappa_2  }  }\norm{ (\covXYZ \covYZX)^{-1}\covXYZ  }_{op} , \sqrt{c_B}  \}, 
   $$
   and the boundedness of referred linear operators are shown in \cref{lem: boundedness of covariance operators}.
   \end{myproof}

\begin{myrem}
There are three maximizers in the proof: $u^*$, $\hat{u}^*$ and $\hat{u}$. 
To clarify the relationship among them, we provide their formal expressions.
The function $u^*$ and $\hat{u}^*$ are the expected maximizer with respect to $f^*$ and $\hat{f}$, respectively:
$$
u^*(y,z) = \mathbb{E}_{X|z}[f^*(X)]-y,
$$
and 
$$
\hat{u}^*(y,z) = \mathbb{E}_{X|z}[\hat{f}(X)]-y.
$$
If we have the full knowledge of the conditional distribution $\prob{X|Z}$, 
the above functions can be computed accurately given $f^*$ and $\hat{f}$.
This will be a future direction when $\prob{X|Z}$ is known to the learners.
However, in many cases, only the samples from $\prob{X|Z}$ can be obtained. 
Hence, an empirical estimator is needed, i.e., 
$$
\hat{u}(y,z) = \hat{\mathbb{E}}_{X|z}[\hat{f}(X)]-y,
$$
where $\hat{\mathbb{E}}_{X|z}$ denotes the empirical estimation for the conditional expectation ${\mathbb{E}}_{X|z}$.
\end{myrem}

\subsection{From finite to infinite dimensional RKHS}

In this subsection, we mainly discuss the infinite-dimensional spaces. 
For finite-dimensional spaces, it suffices to consider their dimensions 
as the Hilbert spaces with the same dimensions are isomorphic.
However, the tool of dimensions is invalid for infinite-dimensional RKHSs. 
We adopt a more powerful tool to describe them, 
more precisely, the decay rate of eigenvalues of a RKHS. 
In subsection \ref{subsec: RKHS}, we introduce the concept of eigen-decomposition of kernels and the decay rate of eigenvalues.
In this subsection, we mainly discuss two types of RKHSs.
\begin{myasp}
   \label{asp: finite}
   The space has finite number of eigenfunctions, 
   i.e., the associated kernel of the space can be expanded in terms of $\tilde{d}$ eigenfunctions.
\end{myasp}
Function classes of this type is $\tilde{d}$-dimensional,
including linear functions, polynomial functions, 
as well as any function class based on the finite number of basis functions. 
Generally, any function space with finite VC-dimension satisfies this condition. 

\begin{myasp}
   \label{asp: infinite}
   The space has countably many eigenfunctions, 
   and eigenvalues satisfy $\mu_i \leq \eta  i^{-2\nu/d}$ for a universal constant $\eta>0$.
\end{myasp}
Function classes of this type is infinite-dimensional,
and the parameter $\nu$ reflects the features of spaces.
This type of scaling covers the case of Sobolev spaces which consists of functions with $\nu$ derivatives.
Besov spaces and Lipschitz-$\nu$ spaces also exhibit this type of eigenvalue decay.
The exponential kernel $k(x,x')=\exp(\gamma\norm{x-x'})$ is a Sobolev-type kernel for $\nu = \frac{d+1}{2}$.
The famous Matern kernel is also of Sobolev-type.

Another tool to capture infinite-dimensional spaces is the \emph{tight metric entropy} \cite{mathLearning}. 
The metric entropy is closely related with packing number of spaces,
whcih intuitively, measures the varieties of functions in the unit ball. 
An $\epsilon$-packing of a metric space $(\mathcal{G},\rho)$ is a collection $\{f_1,\cdots,f_M\}\subset \mathcal{G}$ such that 
$$
\epsilon \leq \rho(f_i , f_j )\leq \gamma \epsilon, \forall i\neq j, 
$$
for some universal constant $\gamma>1$.
The tight $\epsilon$-packing number $\mathfrak{N}( \epsilon ;\mathcal{G}, \rho  ) $ is the cardinality of the largest $\epsilon$-packing of $\mathcal{G}$. 
The metric entropy of a space $\mathcal{H}$ is simply the logarithm of the tight packing number for the unit ball $\mathcal{G}:=\mathcal{B}\subset\mathcal{H}$, i.e., $log \mathfrak{N}( \epsilon ;\mathcal{B}, \rho  ) $, 
which is central to the proof of the lower bounds. 

\citet{optRatesRLSR} point out that for RKHSs, 
the decay rate of eigenvalues is a tighter measure for the complexity of RKHSs than classical entropy assumption.
The relationship between two assumptions is summarized as the following lemma.
The metric entropy of $\tilde{d}$-dimensional spaces are \eqref{log entropy},
and \eqref{poly entropy} is one type of metric entropy for infinite-dimensional spaces.
For the latter case, the dimensions of variables $X$ and $Z$ will be important in analyzing the convergence rate of our methods.

\begin{mylem} [\cite{svm,RKHSminimax}]
   \label{lem: metric entropy}
   For a RKHS satisfying \cref{asp: finite}, then for $\tilde{d} >0$,
   \begin{equation}
      \label{log entropy}
  \log \mathfrak{N}( \epsilon;\mathcal{B}, L^2(\mathbb{P}) ) \simeq  \tilde{d} \log(1/\epsilon), \forall \epsilon \in (0,1). \text{ (logarithmic metric entropy)}.
  \end{equation}
  For a RKHS satisfying \cref{asp: infinite}, then for $\nu>\frac{d}{2}$
  \begin{equation}
   \label{poly entropy}
   log \mathfrak{N}( \epsilon;\mathcal{B}, L^2(\mathbb{P}) ) \simeq   {\epsilon}^{-d/\nu}, \forall \epsilon \in (0,1). \text{ (polynomial metric entropy)}.
   \end{equation}
   
\end{mylem}

\subsection{Lower bound}
\label{sec: lower bound}

Before we discuss the spaces with infinite dimensions, 
we first prove a generalization of \cref{thm: lower bound of finite dimensional spaces}. 
The tool of metric entropy can deal with finite and infinite cases in a unified way.
Let $f(P)$ be some function of $P$, e.g., the mean, the variance or the density of $P$. 
The Tsybakov's minimax theorem shows a powerful application on the proof of lower bounds.

\begin{mythm}[\cite{IntroNonparametricEstimation}]
    \label{thm: Tsybakov minimax}
Let $X_1,\cdots,X_n\sim P_0 \in\mathcal{P}$ be i.i.d. samples.
Assume that $\{P_0,P_1,\cdots,P_N\}\subset \mathcal{P}$ where $N\geq 3$,
and $P_0$ is absolutely continuous with respect to each $P_j$. 
Let $\hat{f}$ be the estimator obtained from samples $X_1,\cdots,X_n$.
Suppose that 
$$
\frac{1}{N}\sum_{j=1}^N \KL(P_j,P_0) \leq \frac{\log N}{16n}.
$$
Then, 
$$
\inf_{\hat{f}} \sup_{P\in \mathcal{P}} \mean{ \rho(\hat{f},f(P) )  }\geq \frac{s}{16},
$$
where 
$$
s = \max_{0\leq j<k\leq N} \rho(f(P_j),f(P_k)  ) .
$$   
\end{mythm}



Consider data $(x_i,y_i)$ satisfying the expression $y_i=f(x_i)+e_i$, 
where $e_i$ follows the normal distribution with zero mean and $\sigma^2$ variance.
To use Tsybakov's result, we need to construct a distribution collection $\mathcal{P} = \{ P_\omega \}$.
Denote $\mathcal{P}$ as the set of distributions of the form $p(x,y|z)=p_\phi(y-f(x))$, 
where $p_\phi$ is the density of normal variables with mean $0$ and variance $\sigma^2$. 
Since $f_\omega$ determines the distribution $P_\omega$, 
we only need to consider the tight metric entropy \cite{mathLearning} of spaces.

\begin{mythm}
    \label{thm: lower bound}
    Consider data $(X_i,Y_i)_{i=1}^n$ following the relationship $Y_i=f(X_i)+E_i$ where $E_i$ is a Gaussian noise, 
    and $\mathcal{H}$ is a RKHS satisfying \cref{asp: finite} (\cref{asp: infinite}), 
    Then for any estimation algorithm $\pi$, 
    there exist a function $f\in\mathcal{H}$ such that 
    $$
    \normX{f-\hat{f}^\pi_n} = \Omega(  \sqrt{\frac{\tilde{d} }{n}})\ \ \ \ (  \normX{f-\hat{f}^\pi_n} = \Omega( n^{  -\frac{\nu}{2\nu+d} } ) )
    $$
    for the estimator $\hat{f}^\pi_n$ obtained by $\pi$ from the data.
\end{mythm}

\begin{myproof}
Let $\mathcal{F}$ be a tight $\epsilon$-packing of a unit ball $\mathcal{B} \subset \mathcal{H}$. 
According to the definition, $|\mathcal{F}| = \mathfrak{N}( \epsilon ;\mathcal{B}, L^2(\mathbb{P}) )$, and 
$$
\int (f-g)^2 d\mathbb{P} > \epsilon^2, \forall f,g \in \mathcal{F}.
$$
We construct $\mathcal{P} = \{ P_\omega =\phi(y-f_\omega(x)): f_\omega \in \mathcal{F} \}$.
Furthermore, the KL divergence of Gaussian variables shows
\begin{equation}
    \label{eq: KL of normal}
\KL(P_\omega,P_\nu) = \frac{1}{2\sigma^2}\int (f_\omega-f_\nu)^2d\mathbb{P} \leq \frac{\gamma^2\epsilon^2}{2\sigma^2}. 
\end{equation}

To apply Tsybakov's theorem, 
we need to have  
$$
\KL(P_\omega,P_\nu) \leq \frac{\log \mathfrak{N}( \epsilon ;\mathcal{B}, L^2(\mathbb{P}) )}{16n}, 
$$
so we can set $\frac{\gamma^2\epsilon^2}{2\sigma^2} = \frac{\log \mathfrak{N}( \epsilon ;\mathcal{B}, L^2(\mathbb{P}) )}{16n}$ to solve the minimax rate $\epsilon$.

Let the distance metric $\rho$ be the norm $\norm{\cdot}_{L^2(\mathbb{P})}$.
Thanks to \cref{thm: Tsybakov minimax},
$$
\inf_{\hat{f}} \sup_{P\in \mathcal{P}} \mean{ d(\hat{f},f(P) )  }\geq \epsilon/16. 
$$
The \cref{lem: metric entropy} shows the metric entropy which the space should satisfy.
For logarithmic metric entropy, $\epsilon \geq C_1 \sigma \sqrt{ \frac{\tilde{d} }{ n}}$ for $n>\frac{\tilde{d} \sigma^2}{8\gamma^2}$; 
for polynomial metric entropy, $\epsilon \geq C_2 \sigma^{ \frac{2\nu}{2\nu+d} } {n^{ - \frac{\nu}{2\nu+d}   }   }$, 
where $C_1>0$ and $C_2>0$ are some constants.
Plugging in the above quantities, we can obtain the lower bound.
\end{myproof}

\begin{myrem}
Many papers \cite{IntroNonparametricEstimation,RKHSminimax,minimax} show that essentially the minimax rate $\epsilon$ can be solved from the Le Cam equation: 
$$
\log \mathfrak{N}( \epsilon;\mathcal{B}, L^2(\mathbb{P}) ) \simeq  n\epsilon^2.  
$$
\end{myrem}
     
We establish the lower bound for two special cases of metric entropy. 
The reproducing kernel Hilbert spaces with logarithmic metric entropy only have $\tilde{d}$ nonzero eigenvalues, 
so the decay rate can be viewed as the ``fastest'' to some degree. 
On the contrary, the eigenvalues of the spaces with polynomial metric entropy decay the ``slowest'' to some extent. 
Actually, the decay rate of many RKHSs is between above two, such as exponential decay rate. 

For RKHS satisfying \eqref{log entropy}, it can be regarded as an extreme case when $\nu\to\infty$. 
At high-level, finite nonzero eigenvalues mean these eigenvalues converge to zero arbitrary fast. 
Hence, the convergence rate should be a limit case of these with rate $j^{-2\nu}$ for large $\nu$. 
Let $\nu\to\infty$ we obtain a rate $\Omega(\sqrt{\frac{1}{n}  })$. 
However, this reasoning ignores the effect of $\tilde{d}$ and the optimal order of lower bound should be $\Omega(\sqrt{\frac{\tilde{d} }{n}  })$ from \cref{thm: lower bound}.

The proof can be easily extended to truncated normal variables 
because the upper bound of KL-divergence of two truncated normal variables in \cref{eq: KL of normal} scales a density normalized constant \citep{truncatednormal}. 
Indeed, a normal distribution behaves very much like a bounded one due to its thin tails.

\subsection{Oracle inequality for infinite-dimensional RKHSs}

For a regularized least squares regression,
denote the estimator solved from a general penalty term:
\begin{equation}
   \label{eq: f solved in a general form}
\hat{f} \in \argmin_{f\in\mathcal{H}} \lambda(f) + \hat{\mathcal{R}}(f).
\end{equation}
For the problem \eqref{eq: f solved in quadratic loss}, it is a special form with $\lambda(f) = \lambda \normH{f}^2$.
To achieve an optimal rate, we need the main theorem in \cite{optRatesRLSA}.
We restate it as a corollary to make it consistent with our notations.

\begin{mycol}[A simplified verson of the main theorem in \cite{optRatesRLSA}]
\label{thm: rate of OLS}
Let the RKHS $\mathcal{H}$ satisfy the \cref{asp: infinite} 
and the corresponding kernel $k$ satisfy \cref{asp: continuity}.
Then, there exists some constants $c$, 
such that for all $ \tau > 0$ and 
$\lambda_i  \simeq  ( \log n)^{ \frac{\nu}{2\nu+d} }   n^{-\frac{\nu}{2\nu+d} }  $ for $i=1,2$,
the learning method described by \eqref{eq: f solved in a general form} satisfy 
$$
\mathcal{R}(\hat{f}) - \mathcal{R}({f}^*) \leq  c\tau ( \log n)^{ \frac{2\nu}{2\nu+d} }   n^{-\frac{2\nu}{2\nu+d} }
$$
with probability not less than $1-e^{-\tau}$.
\end{mycol}

\begin{myrem}
   \citet{regularizationKernel} consider a more precise penalty term, 
   with penalty terms of order 
   $\lambda(f) \simeq  (\tau+ \log n + \log\log \normH{f} ) (\log n)^{ \frac{4\nu}{2\nu+d} }   n^{-\frac{2\nu}{2\nu+d}} \normH{f}^{  \frac{2d}{2\nu +d} }  $ .
   They also obtain the optimal rate $\mathcal{O}( ( \log(1/\delta) + \log n)  (\log n)^{ \frac{4\nu}{2\nu+d} }   n^{-\frac{2\nu}{2\nu+d}}  )$.
   However, the penalty term is too complicated and there does not exist a closed-form solution. 
   We leave the analysis of duelIV with this penalty as the future work. 
   
\end{myrem}

Now we can establish the following oracle inequality for the solution obtained from dualIV with quadratic penalty terms.

\begin{mythm}[Oracle Inequality]
   \label{thm: oracle inequality}
   Let the RKHS $\mathcal{H}$ and $\mathcal{U}$ satisfy the assumption \eqref{log entropy},
   and $k$ and $l$ are corresponding kernels on compact sets $\mathcal{X} \subset \mathbb{R}^d$ and $\mathcal{Y}\times \mathcal{Z}\subset \mathbb{R}^d$, respectively.
   Suppose that \cref{asp: realizability,asp: invertibility,asp: continuity} hold.
   Then, there exists a constant $c$ 
   which depends on the true structural function $f^*$ and spaces $\mathcal{H},\mathcal{U}$,
   such that for sufficiently large $n$, 
   the convergence rate of $\hat{f}$ obtained from dualIV with penalty terms 
   $\lambda_i  \simeq  ( \log n)^{ \frac{\nu}{2\nu+d} }   n^{-\frac{\nu}{2\nu+d} }  $ for $i=1,2$,
   satisfies 
   $$
   \norm{\hat{f}-f^*}_{L^2(\mathbb{P}_{X})} \leq c \sqrt{\tau } (\log n)^{ \frac{2\nu}{2\nu+d} }  n^{-\frac{\nu}{2\nu+d}}  
   $$
   with probability not less than $1-2e^{-\tau}$.
\end{mythm}

\begin{myproof}
The proof is the same as the steps taken in the proof of \cref{thm: oracle inequality for finite rank}. 
The only difference is the order of convergence rates. 
During the proof, we need to replace the order $\mathcal{O}(\sqrt{\tilde{d}/n})$ with $\mathcal{O}( \sqrt{\tau  } (\log n)^{ \frac{2\nu}{2\nu+d} }  n^{-\frac{\nu}{2\nu+d}}  )$, 
because we replace the \cref{thm: rate of OLS for finite rank kernel} with \cref{thm: rate of OLS} during the proof.

\end{myproof}

\begin{myrem}
Compared with \cref{thm: oracle inequality for finite rank}, 
this theorem is related with the dimensions of input variables. 
Specifically, it requires that the dimension of $Z$ is less than or equal to $d-1$, 
where $d$ is the dimension of $X$.
This fact demonstrates the essential difference between spaces with finite and infinite dimensions.
\end{myrem}

\section{Regret Analysis}

\subsection{Regret upper bound}

\begin{mythm}
    Suppose that \cref{asp: realizability,asp: invertibility,asp: continuity} hold in kernelized contextual bandit settings.
    Moreover, the epoch schedule is set to be $\tau_m =  2^m$, 
    and the tuning parameters are properly selected to match the constant in \cref{thm: oracle inequality for finite rank}.
    Then, the expected regret $Reg(T)$ of \cref{alg: DIV-BLS} is at most
    $$
     \mathcal{O}(2\sqrt{KT \tilde{d} \log (2\log^2(T) /\delta)} +  \sqrt{8T \log(2/\delta)} )
    $$
    with probability at least $1-\delta$.
\end{mythm}
    
    \begin{myproof}
        Thanks to the the epoch learning strategy and \cref{thm: oracle inequality for finite rank}, 
        the following event 
        $$
        \mathcal{E}_m=\left\{ \forall \tau_{m-2}+1  \leq t \leq \tau_{m-1}, \meanX{ \hat{f}_m(x_t) - {f}^*(x_t) } \leq \frac{K}{4\gamma_m^2} |\mathcal{F}_{t-1}   \right\}
        $$
        happens with probability at least $1- \frac{\delta}{2m^2}$. 
        Therefore, by a union bound, the event $\bigcap_{m}  \mathcal{E}_m  $ holds with probability at least $1-\delta/2$, 
        as 
        $$
        1 - \sum_{m=2} \frac{\delta}{2m^2} = 1- \frac{\pi^2 -6 }{12} \delta > 1- \delta/2.
        $$
        Assume the above event holds. 
        Then by the lemma 10 in \cite{fasterCB}, 
        the expected regret after $T$ rounds is at most 
        $$
        \sum_{t=\tau_1+1}^T 8K/\gamma_{m(t)} +\tau_1 + \sqrt{8T \log(2/\delta) },
        $$
        where $m(T)$ is the total number of epochs at the round $T$.
        We sum over all $\gamma_m$ and can obtain the regret upper bound.
        $$
        \begin{array}{rcl}
        Reg(T) &\leq& \sum_{t=\tau_1+1}^T 8K/\gamma_{m(t)} +\tau_1 + \sqrt{8T \log(2/\delta) } \\
        &\leq& \sum_{m=2}^{m(T)} \sqrt{  K\tilde{d} (\tau_{m-1}-\tau_{m-2}) \log( 2m^2/\delta)  )    /\eta    } + 2+ \sqrt{8T \log(2/\delta) } \\
        &<&  \sum_{m=2}^{m(T)} \sqrt{  K\tilde{d} 2^{m-2} \log( 2m^2/\delta)      /\eta    } + 2+ \sqrt{8T \log(2/\delta) } \\
        &<&  \int_{2}^{m(T) } \sqrt{  K\tilde{d} 2^{m-2} \log( 2m^2/\delta)      /\eta    } dm + 2+ \sqrt{8T \log(2/\delta) } \\
        &<&  \sqrt{  K\tilde{d} \log( 2\log^2 (T) /\delta)      /\eta }  \int_{2}^{m(T)} 2^{m/2-1}  dm + 2+ \sqrt{8T \log(2/\delta) } \\
        &<&  2\sqrt{  KT\tilde{d} \log( 2\log^2 (T) /\delta)      /\eta } + 1 + \sqrt{8T \log(2/\delta) }  \\
        \end{array}
        $$
        
 \end{myproof}

\subsection{Regret lower bound}

\begin{mythm}
Assume that $\mathcal{H}$ is $\tilde{d}$-dimensional. 
Moreover, $K\leq 2^{\tilde{d}/2}$ and $T\geq \tilde{d}(\log K)^{1+\epsilon} $ for any small constant $\epsilon>0$.
For any algorithm $\pi$, 
there exists an bandit instance with a reward function $f\in\mathcal{H}$ such that 
$$
Reg(T)\geq \Omega \left( \sqrt{ \tilde{d} T \log K \log(T/\tilde{d})  } \right).
$$
\end{mythm}
\begin{myproof}
Regard $\phi(x)$ as the new context in $\mathcal{H}$.
Then due to the representor's theorem of RKHSs, there exists a $\tilde{d}$-dimensional element $\theta\in \mathcal{H}$ such that 
$$
f(x) = \innerH{\theta}{\phi(x)}.
$$
Hence, the kernelized contextual bandit can be reduced to a linear bandit in the new feature space. 
Apply the lower bound in \cite{minimaxOptimalLin} with the context dimension to be $\tilde{d}$. 
Then we obtain an instance with parameter $\theta^*$ such that 
$$
Reg(T)\geq \Omega( \sqrt{ \tilde{d} T \log K \log(T/\tilde{d})   }).
$$
We conclude the proof by setting the reward function to be $\innerH{\theta^*}{\phi(x)}$.

\end{myproof}

\begin{mythm}
Suppose $K>1$ and $\tilde{d}>1$. 
Consider the function space $\mathcal{H}$ of the form \eqref{eq: multiple reward function space}.
Then for any algorithm $\pi$, 
there exists a reward function $f \in \mathcal{H} $ such that 
$$
Reg(T)\geq \Omega\left(  \sqrt{\tilde{d}KT}  \right).
$$
\end{mythm}

\begin{myproof}
Without loss of generally, we assume that $[K]$ is the whole action set.
We construct the following $[(m-1)(K-1)+1]$-armed bandit,
where each $\theta_{ij} b_j(x)$ is an independent arm. 
\begin{table}[hbtp]
    \begin{center}
    \begin{tabular}{c|cccc}
    arm/basis & $b_1(x)$ & $b_2(x)$            & $\cdots$ & $b_m(x)$            \\ \hline
    1            & $mb_1(x)$ & 0                   & 0        & 0                   \\
    $\vdots$     & 0        & $\theta_{22} b_2(x)$ & $\cdots$ & $\theta_{2m} b_m(x)$ \\
    $K$          & 0        & $\theta_{K2} b_2(x)$ & $\cdots$ & $\theta_{Km} b_m(x)$
    \end{tabular}
    \end{center}
\end{table}
Moreover, we assume all rewards are nonnegative, 
because we can add a large constant to each term without changing proof. 
The coefficients will be specified later. 
Hence in our setting, when the learner pulls an arm in $[K]$, it collects a row of rewards in the above table.
We define the following reward functions for kernelized contextual bandits.
$$
f_1(x) = mb_1(x), f_a(x) =  \sum_{j=2}^m \theta_{aj} b_j(x), a=2,3,\cdots,K.
$$

Let $f_1(x)$ be the optimal arm for $x\in\mathcal{X}$ in kernelized contextual bandit setting, 
and $b_1(x)$ be optimal in $[(m-1)(K-1)+1]$-armed bandit setting.
Based on the algorithm $\pi$, we construct the policy $\pi'$ for the MAB problem. 
At round $t$, the algorithm $ \pi$ pull the arm $a_t$. 
If $a_t =1$, then $\pi'$ chooses the arm $b_1(x)$. 
If not, the algorithm $\pi'$ randomly choose a column $\pi'(t)$ in the row $a_t$ of the above table, 
and thus obtains the reward $\theta_{a_t, \pi'(t)}b_{\pi'(t)}$. 
In other words, $\pi'$ pulls the arm $\theta_{a_t, \pi'(t)}b_{\pi'(t)}(x) $.
For notation brevity, let $\theta_{11}=m, \theta_{1i}=0$ for $i=2,\cdots,m$.
Then 
$$
\begin{array}{rcl}
    Reg(T)  &=&  \sum_{t=1}^T  \mathbb{E}[   m b_1(x_t) -    \sum_{j=1}^m  \theta_{a_t,j} b_j(x_t)    ]\\
    &=&  \sum_{t=1}^T   \sum_{j=1}^m  \mathbb{E}[   b_1(x_t) -    \theta_{a_t, j } b_j(x_t)    ]\\
    &=&  \sum_{t=1}^T   \sum_{j=1}^m  \mathbb{E}[   b_1(x_t) -    \theta_{a_t, j } b_j(x_t)    ]  \ind{a_t\neq 1} \\
    &\geq&  \sum_{t=1}^T   \mathbb{E}[   b_1(x_t) -    \theta_{a_t, \pi'(t) } b_{\pi'(t)}(x_t)    ] \ind{a_t\neq 1}  \\
\end{array}
$$
The RHS can be considered as the expected regret of a $[(m-1)(K-1)+1]$-armed bandit. 
Due to the theorem of lower bound in \cite{banditBook}, 
the expected regret is $\Omega(\sqrt{mKT})$ for certain instance. 
We then can select appropriate positive values of $\theta_{ij}$ to satisfy the conditions of the lower bound theorem. 
Finally, we conclude the proof by plugging in the above lower bound.

\end{myproof}

\begin{mythm} 
Suppose $K>1$ and $\tilde{d}>1$. 
Consider the function space $\mathcal{H}$ of the form \eqref{eq: multiple reward function space}.
Then for any algorithm $\pi$ and $\delta \in (0,1)$ satisfying 
$$
T\delta \leq \sqrt{ T  K\tilde{d} \log \Big( \frac{1}{4\delta} \Big)  },
$$
there exists a reward function $f \in \mathcal{H} $ such that 
$$
Reg(T)\geq    \frac{1}{4} \min  \Big\{  T,  \sqrt{\tilde{d}KT  \log \Big(\frac{1}{4\delta} \Big) }  \Big\}  
$$
with probability at least $1-\delta$.
\end{mythm}
\begin{myproof}
Regard the kernelized contextual bandit as a $K\tilde{d}$-armed bandit, 
and only consider the cases when $\norm{f}_{\infty}\leq 1$.
Assume that the conclusion does not hold for an algorithm $\pi$ and let $\delta$ satisfy the condition. 
Then for any bandit instance with $f\in\mathcal{H}$, 
the expected regret of $\pi$ is bounded by 
$$
Reg(T)\leq T\delta +\sqrt{ T  K\tilde{d} \log \Big( \frac{1}{4\delta} \Big)  } \leq 2 \sqrt{ T  K\tilde{d} \log \Big( \frac{1}{4\delta} \Big)  }.
$$
Then $\pi$ satisfy the condition of Theorem 17.1 in \cite{banditBook} with $B = 2\sqrt{ \log \Big( \frac{1}{4\delta} \Big)  }$, 
which implies the conclusion holds for some instance with $\norm{f}_{\infty}\leq 1$,
contradicting the assumption.

\end{myproof}

\subsection{Extensions}

We can extend the algorithm DIV-ELS to infinite-dimension cases. 
We generalize \cref{alg: DIV-BLS} to \cref{alg: DIV-BLS for infinite case}.

\begin{algorithm}
	\renewcommand{\algorithmicrequire}{\textbf{Input:}}
	\renewcommand{\algorithmicensure}{\textbf{Output:}}
	\caption{DualIV with Epoch Learning Strategy for Infinite-dimensional RKHSs}
	\label{alg: DIV-BLS for infinite case}
    \begin{algorithmic}[1]
        \Require epoch schedule $0=\tau_0<\tau_1<\tau_2<\cdots$, confidence parameter $\delta$, kernel functions $k,l$, tuning parameter $\eta,\eta_1,\eta_2$
        \State determine the parameter $\nu$ and the dimension $d$ according to kernels $k,l$ and the data, respectively 
        \For{ epoch $m = 1,2,\cdots,$ }
           \State collect (only) the data in epoch $m-1$, e.g., 
              $$
               \mathcal{D}_{m-1} = \{ (a_{\tau_{m-2}+1  }, c_{\tau_{m-2}+1  },y_{\tau_{m-2}+1  }, z_{\tau_{m-2}+1  }   ),\cdots,  (a_{\tau_{m-1}  }, c_{\tau_{m-1}  },y_{\tau_{m-1}  }, z_{\tau_{m-1}  }   )   \}
              $$
           \State let regularization parameters $\lambda_i = \eta_i   ( \log |\mathcal{D}_{m-1} | )^{ \frac{\nu}{2\nu+d} }  |\mathcal{D}_{m-1} |^{-\frac{\nu}{2\nu+d} } $ for $i=1,2$
           \State implement dualIV with input $\lambda_1,\lambda_2$, $k,l$ and $\mathcal{D}_{m-1}$, 
                  and then obtain $\hat{f}_m$ (for epoch $1$, $\hat{f}_1=0$)
           \State compute $\gamma_m =  \sqrt{  \frac{\eta K   }{  \log (2m^2/\delta)  }  }  (\log |\mathcal{D}_{m-1} |)^{ -\frac{\nu}{2\nu+d}} |\mathcal{D}_{m-1} |^{\frac{\nu}{2\nu+d} }  $ (for epoch $1$, $\gamma_1 = 1$)
              \For{ round $t = \tau_{m-1}+1,\cdots, \tau_{m}$ }
              \State observe the context $c_t$ and the instrumental variable $z_t$
              \State  compute $\hat{f}_m(c_t,a)$ for each action $a\in\mathcal{A}$ and the following probabilities
              $$
              p_t(a) = \left\{
              \begin{array}{lcl}
                  \frac{1}{K+\gamma_m (\hat{f}_m (x_t, \hat{a}_t )- \hat{f}_m (x_t, a ) ) }, & & \text{ for all } a\neq \hat{a}_t \\
                  1- \sum_{a\neq \hat{a}_t }  p_t(a), & &\text{ for } a = \hat{a}_t .
              \end{array}
              \right.
              $$
              where $\hat{a}_t =\max_{a\in\mathcal{A}} \hat{f}_m(x_t,a)   $.
              \State sample $a_t \sim p_t(\cdot)$ and observe reward $y_t$
              \EndFor
           \EndFor
   \end{algorithmic}  
\end{algorithm}

The regret of \cref{alg: DIV-BLS for infinite case} can be analyzed in a similar way. 
We omit the proof of the following theorem as it only simply repeats the steps of \cref{thm: regret upper bound}.
\begin{mythm}
    Suppose that assumptions \eqref{asp: realizability}, \eqref{asp: invertibility}, \eqref{asp: continuity} and \eqref{asp: infinite} hold in kernelized contextual bandit settings.
    Moreover, the epoch schedule is set to be $\tau_m =  2^m$, 
    and the tuning parameters are properly selected to match the constant in \cref{thm: oracle inequality}.
    Then, the expected regret $Reg(T)$ of algorithm \cref{alg: DIV-BLS for infinite case} is at most
    $$
     \mathcal{O}( \sqrt{K \log( 2\log^2 (T) /\delta)} (\log T)^{ \frac{2\nu}{2\nu+d} +\frac{1}{2} } T^{\frac{ \nu+d}{2\nu+d}}  +\sqrt{8T \log(2/\delta) } )
    $$
    with probability at least $1-\delta$.
\end{mythm}

\citet{LowerBoundInfKernel} prove that the lower bound of kernelized bandits with Mat{\'e}rn-$\nu$ kernel is $\Omega(T^{ \frac{\nu+d}{ 2\nu +d} })$, 
but this result is not general enough,
because they focus on kernels of specific type. 
In \cite{CBwithRegressionOracles}, under the assumption of tensorization (i.e., the function spaces of the form \eqref{eq: multiple reward function space}),
the lower bound for RKHSs satisfying \eqref{poly entropy} is $\Omega(T^{ \frac{\nu+d}{ 2\nu +d} })$. 
Therefore, the regret upper bound is rate-optimal up to logarithmic terms.

\end{document}